\documentclass{article}

\usepackage[table,svgnames]{xcolor}
\usepackage[accepted]{icml2023}

\usepackage{amsmath,amsfonts,bm}

\def\eqref#1{equation~\ref{#1}}

\def\1{\bm{1}}

\DeclareMathAlphabet{\mathsfit}{\encodingdefault}{\sfdefault}{m}{sl}
\SetMathAlphabet{\mathsfit}{bold}{\encodingdefault}{\sfdefault}{bx}{n}

\def\gD{{\mathcal{D}}}

\def\gL{{\mathcal{L}}}

\def\gP{{\mathcal{P}}}

\def\gS{{\mathcal{S}}}

\def\gU{{\mathcal{U}}}

\def\gX{{\mathcal{X}}}
\def\gY{{\mathcal{Y}}}
\def\gZ{{\mathcal{Z}}}

\newcommand{\E}{\mathop{\mathbb{E}}}

\newcommand{\R}{\mathbb{R}}

\DeclareMathOperator*{\argmax}{arg\,max}
\DeclareMathOperator*{\argmin}{arg\,min}

\DeclareMathOperator*{\proj}{proj}

\usepackage[utf8]{inputenc}
\usepackage[T1]{fontenc}
\usepackage{textcomp}
\usepackage{float}
\usepackage[normalem]{ulem}

\usepackage{graphicx}
\usepackage{booktabs}
\usepackage{diagbox}
\usepackage{numprint}
\usepackage{wrapfig}
\usepackage{multirow}

\usepackage{listings}
\definecolor{linkcolor}{RGB}{74, 102, 146}
\newcommand{\cellhi}{\cellcolor{RoyalBlue!15}}
\usepackage[colorlinks=true,allcolors=linkcolor,pageanchor=true,plainpages=false,pdfpagelabels,bookmarks,bookmarksnumbered, backref=page]{hyperref}
\usepackage{transparent}

\renewcommand*{\backref}[1]{}
\renewcommand*{\backrefalt}[4]{\ifcase #1 Not cited.%
  \or Cited on page~#2.%
  \else Cited on pages #2.%
  \fi%
}

\usepackage{graphicx}
\usepackage{hyperref}
\usepackage{url}

\usepackage[edges]{forest}

\usepackage{algpseudocode}
\algnewcommand{\LeftCommentX}[1]{\Statex \(\triangleright\) #1}
\algnewcommand{\LeftComment}[1]{\State \(\triangleright\) #1}

\usepackage{tikz}
\usetikzlibrary{arrows,backgrounds,bayesnet,calc,matrix}

\newcommand{\cblock}[3]{
  \hspace{-1.5mm}
  \begin{tikzpicture}[node/.style={square, minimum size=10mm, thick, line width=0pt}]
    \node[fill={rgb,255:red,#1;green,#2;blue,#3}] () [] {};
  \end{tikzpicture}%
}

\newcommand{\circlegend}[2]{
  \hspace{-1.5mm}
  \begin{tikzpicture}[node/.style={square, minimum size=10mm, thick, line width=0pt}]
    \node[circle,fill={#1},draw=black,thick] (block) [] {};
    \node[right=0mm of block,yshift=-.3mm,minimum height=6mm] {#2};
  \end{tikzpicture}%
}

\usepackage{algorithm, algpseudocode}

\usepackage[nameinlink]{cleveref}
\Crefname{equation}{Eq.}{Eqs.}
\Crefname{section}{Sect.}{Sects.}
\Crefname{appendix}{App.}{Apps.}
\Crefname{definition}{Def.}{Defs.}
\Crefname{proposition}{Prop.}{Props.}

\usepackage{xspace}
\newcommand{\eg}{e.g.\xspace}
\newcommand{\ie}{i.e.\xspace}

\usepackage{mathtools}
\newcommand{\defeq}{\vcentcolon=}

\usepackage{xspace}
\definecolor{bdacolor}{RGB}{168, 141, 201}

\usepackage[misc]{ifsym} %
\usepackage[symbol]{footmisc}

\icmltitlerunning{Meta Optimal Transport}

\begin{document}

\twocolumn[
\icmltitle{Meta Optimal Transport}

\begin{icmlauthorlist}
\icmlauthor{Brandon Amos}{meta}
\icmlauthor{Giulia Luise}{msr}
\icmlauthor{Samuel Cohen}{meta,ucl,fairgen}
\icmlauthor{Ievgen Redko}{huawei,aalto}
\end{icmlauthorlist}

\icmlaffiliation{meta}{Meta AI}
\icmlaffiliation{msr}{Microsoft Research}
\icmlaffiliation{ucl}{University College London}
\icmlaffiliation{aalto}{Aalto University}
\icmlaffiliation{fairgen}{Fairgen}
\icmlaffiliation{huawei}{Noah's Ark Lab, Huawei}

\icmlcorrespondingauthor{Brandon Amos}{bda@meta.com}

\icmlkeywords{optimal transport, amortized optimization}

\vskip 0.3in
]

\printAffiliationsAndNotice{} %

\begin{abstract}
  We study the use of amortized optimization to predict optimal transport (OT) maps from the input measures, which we call \emph{Meta OT}. This helps repeatedly solve similar OT problems between different measures by leveraging the knowledge and information present from past problems to rapidly predict and solve new problems. Otherwise, standard methods ignore the knowledge of the past solutions and suboptimally re-solve each problem from scratch. We instantiate Meta OT models in discrete and continuous settings between grayscale images, spherical data, classification labels, and color palettes and use them to improve the computational time of standard OT solvers. Our source code is available at \url{http://github.com/facebookresearch/meta-ot}.
\end{abstract}

\section{Introduction}
Optimal transportation \citep{villani2009optimal,ambrosio2003lecture,santambrogio2015optimal,peyre2019computational,merigot2021optimal} is thriving in domains including
economics \citep{galichon2016optimal},
reinforcement learning \citep{dadashi2021primal,fickinger2021cross},
style transfer \citep{kolkin2019style},
generative modeling \citep{arjovsky2017wasserstein,seguy2018large,huang2020convex,rout2021generative},
geometry \citep{solomon2015convolutional,cohen2021riemannian},
domain adaptation \citep{courty2017joint,redko2019optimal},
signal processing \citep{kolouri2017optimal},
fairness \citep{jiang2020wasserstein}, and cell reprogramming \citep{SCHIEBINGER2019928}.
These settings couple two measures $(\alpha,\beta)$ supported
on domains $(\gX, \gY)$ by solving a transport
optimization problem such as the \emph{primal Kantorovich problem}
defined by
\begin{equation}
  \pi^\star(\alpha, \beta, c)\in \argmin_{\pi\in\gU(\alpha, \beta)}
    \int_{\gX\times \gY} c(x,y){\rm d}\pi(x,y),
  \label{eq:kantorovich-primal}
\end{equation}
where the \emph{optimal coupling} $\pi^\star$
is a joint distribution over the  product space,
$\gU(\alpha, \beta)$ is the set of admissible couplings
between $\alpha$ and $\beta$, and
$c: \gX\times\gY \rightarrow \R$ is the \emph{ground cost}, that represents a notion of distance between elements in $\mathcal{X}$ and elements in $\mathcal{Y}$.

\textbf{Challenges.}
Unfortunately, solving \cref{eq:kantorovich-primal} \emph{once} is
computationally expensive between general measures  and
computationally cheaper alternatives are an active research topic:
\emph{Entropic optimal transport} \citep{cuturi2013sinkhorn}
smooths the transport problem with an entropy penalty, and
\emph{sliced distances}
\citep{kolouri2016sliced,kolouri2018sliced,kolouri2019generalized,deshpande2019max}
solve OT between 1-dimensional projections of the measures,
where \cref{eq:kantorovich-primal} can be solved easily.

When an optimal transport method is deployed
in practice, \cref{eq:kantorovich-primal} is not just
solved once, but is \emph{repeatedly} solved for new
scenarios between different input measures $(\alpha, \beta)$.
For example, the measures could be representations of images
we care about optimally transporting between and in deployment
we would receive a stream of new images to couple.
Repeatedly solving optimal transport problems also comes up in
the context of comparing seismic signals \citep{engquist2013application}
and in single-cell perturbations \citep{bunne2021learning,bunne2022proximal,bunne2022supervised}.
Standard optimal transport solvers deployed in this
setting re-solve the optimization problems from
scratch and ignore the shared structure and
information present between different coupling problems.

\textbf{Overview.}
We study the use of amortized optimization and machine learning
methods to rapidly solve multiple optimal transport problems
and predict the solution from the input measures $(\alpha, \beta)$.
This setting involves learning a \emph{meta} model
to predict the solution to the optimal transport problem,
which we will refer to as \emph{Meta Optimal Transport}.
We learn Meta OT models to predict the solutions to optimal
transport problems and significantly improve the computational
time and number of iterations needed to solve
\cref{eq:kantorovich-primal}.

\textbf{Settings that are not Meta OT.}
Meta OT is not useful in settings that do \emph{not}
\emph{repeatedly} solve OT problems, \eg
1) generative modeling settings, such
as \citet{arjovsky2017wasserstein}, that estimate the
OT distance between the data and model distributions,
and 2) out-of-sample settings \citep{seguy2018large,perrot2016mapping}
that couple measures and then extrapolate the map to
larger measures.

\section{Preliminaries and background}\label{sec:preliminaries_andback}
\subsection{Entropic OT between discrete measures}
\label{sec:prelim:discrete}

We review foundations of OT, following
the notation of \citet{peyre2019computational} in most places.
The discrete setting often favors the entropic
regularized version since it can be computed efficiently
and in a parallelized way using the Sinkhorn algorithm.
While the primal Kantorovich formulation in \cref{eq:kantorovich-primal}
provides an intuitive problem description, OT problems
are rarely solved directly in this form due to the high-dimensionality
of the couplings $\pi$ and the difficulty of satisfying the
coupling constraints $\gU(\alpha, \beta)$.
Instead, most computational OT solvers use the
\emph{dual} of \cref{eq:kantorovich-primal},
which we build our Meta OT solvers on top of.

Let $\alpha\defeq\sum_{i=1}^m a_i\delta_{x_i}$ and
$\beta\defeq\sum_{i=1}^n b_i\delta_{y_i}$
be
\emph{discrete} measures,
where $\delta_z$ is a Dirac at point $z$
and $a\in\Delta_{m-1}$ and $b\in\Delta_{n-1}$ are
in the \emph{probability simplex} defined by
\begin{equation}
  \label{eq:simplex}
  \Delta_{k-1}\defeq\{x\in \R^k : x\geq 0\; {\rm and}\; \sum_i x_i=1\}.
  \footnotesize
\end{equation}

\vspace{-4mm}
\textbf{Discrete OT.}
\Cref{eq:kantorovich-primal} becomes the \emph{linear program} \\[-3mm]
\begin{equation}
  P^\star(\alpha,\beta,c)\in \argmin_{P\in U(a,b)} \langle C, P \rangle \qquad
  \label{eq:discrete-primal}
\end{equation}\\[-3mm]
where
$U(a, b)\defeq \small \{P\in\R_+^{n\times m} : P 1_m = a,\quad P^\top 1_n = b\}$, $P$ is a \emph{coupling matrix},
$P^\star(\alpha, \beta)$ is the \emph{optimal} coupling,
and the \emph{cost} can be discretized as a matrix $C\in\R^{m\times n}$ with entries
$C_{i,j}\defeq c(x_i, y_j)$, and
$\langle C, P\rangle \defeq \sum_{i,j}C_{i,j}P_{i,j}$,

\textbf{Entropic OT.}
The linear program above can be regularized adding an
entropy term to smooth the objective as in
\citet{cominetti1994asymptotic,cuturi2013sinkhorn},
resulting in:
\begin{equation}
  P^\star(\alpha,\beta,c,\epsilon)\in \argmin_{P\in U(a,b)} \langle C, P \rangle - \epsilon H(P)
  \label{eq:entropic-primal}
\end{equation}
where $H(P)\defeq -\sum_{i,j} P_{i,j}(\log(P_{i,j})-1)$
is the discrete entropy of a coupling matrix $P$.

\textbf{Entropic OT dual.}
As presented in \citet[Prop.~4.4]{peyre2019computational}, setting
$K\in\R^{m\times n}$ to the \emph{Gibbs kernel}
$K_{i,j}\defeq \exp\{-C_{i,j}/\epsilon\}$,
the dual of \cref{eq:entropic-primal} is
\begin{equation}
  f^\star,g^\star\in\argmax_{f\in\R^n,g\in\R^m} \langle f, a\rangle + \langle g, b \rangle
    - \epsilon e^{f/\epsilon} K e^{g/\epsilon}
  \label{eq:entropic-dual}
\end{equation}
where the \emph{dual variables} or \emph{potentials}
$f\in\R^n$ and $g\in\R^m$ are associated, respectively, with the
marginal constraints $P1_m=a$ and $P^\top 1_n=b$.
We omit the dependencies of the duals on the context,
\eg $f^\star$ is shorthand for $f^\star(\alpha, \beta, c, \epsilon)$.

\textbf{Recovering the primal solution from the duals.}
Given optimal duals $f^\star$, $g^\star$ that solve
\cref{eq:entropic-dual} the optimal coupling
$P^\star$ to the primal problem in \cref{eq:entropic-primal}
can be obtained by
\begin{equation}
  P_{i,j}^\star(\alpha, \beta, c, \epsilon)\defeq \exp\{f_i^\star/\epsilon\}K_{i,j}\exp\{g_j^\star/\epsilon\}.
  \qquad
  \label{eq:entropic-duals-to-primal}
\end{equation}

\textbf{The Sinkhorn algorithm}.
\Cref{alg:sinkhorn} summarizes the log-space version,
which takes closed-form block coordinate ascent updates
on \cref{eq:entropic-dual}
obtained from the first-order optimality conditions
\citep[Remark~4.21]{peyre2019computational}.
We will fine-tune Meta OT predictions with Sinkhorn.

\textbf{Computing the error.}
Standard implementations of the Sinkhorn algorithm,
such as \citet{flamary2021pot,cuturi2022optimal},
measure the error of a candidate dual solution $(f,g)$ by
computing the deviation from the marginals:
\begin{equation}
  {\rm err}(f,g; \alpha, \beta, c)\defeq\|P1_m-a\|_1 + \|P^\top1_n-b\|_1,
  \qquad
  \label{eq:sinkhorn-err}
\end{equation}
where $P$ is computed from  \cref{eq:entropic-duals-to-primal}.

\textbf{Mapping between the duals.}
The first-order optimality conditions of \cref{eq:entropic-dual}
also provide an equivalence between the optimal dual potentials
that we will make use of
\begin{equation}
  g(f; b, c) \defeq \epsilon \log b - \epsilon\log\left(
    K^\top \exp\{f/\epsilon\}\right).
  \label{eq:f-to-g}
\end{equation}

\subsection{OT between continuous (Euclidean) measures}
\label{sec:prelim:continuous}

\begin{figure}[t]
\vspace*{-3mm}
\begin{algorithm}[H]
\caption{\footnotesize Sinkhorn($\alpha, \beta, c, \epsilon, f_0=0$)}
\begin{algorithmic}
    \footnotesize
    \For {iteration $i$ = 1 to $N$}
    \State $g_{i} \leftarrow \epsilon \log b - \epsilon\log\left(
    K^\top \exp\{f_{i-1}/\epsilon\}\right)$
    \State $f_{i} \leftarrow \epsilon \log a - \epsilon\log\left(
    K \exp\{g_i/\epsilon\}\right)$
    \EndFor
    \State Compute $P_N$ from $f_N,g_N$ using \cref{eq:entropic-duals-to-primal}
    \State\Return $P_N\approx P^\star$
\end{algorithmic}
\label{alg:sinkhorn}
\end{algorithm}
\vspace*{-6mm}
\begin{algorithm}[H]
\caption{\footnotesize W2GN($\alpha, \beta, \varphi_0$)}
\begin{algorithmic}
\footnotesize
\For {iteration $i$ = 1 to $N$}
\State Sample from $(\alpha,\beta)$ and estimate $\gL(\varphi_{i-1})$
  (\cref{eq:w2gn-loss})
\State Update $\varphi_i$ with approximation to $\nabla_\varphi \gL(\varphi_{i-1})$
\EndFor
\State \Return $T_N(\cdot)\defeq\nabla_x \psi_{\varphi_N}(\cdot) \approx T^\star(\cdot)$
\end{algorithmic}
\label{alg:w2gn}
\end{algorithm}
\vspace*{-8mm}
\end{figure}

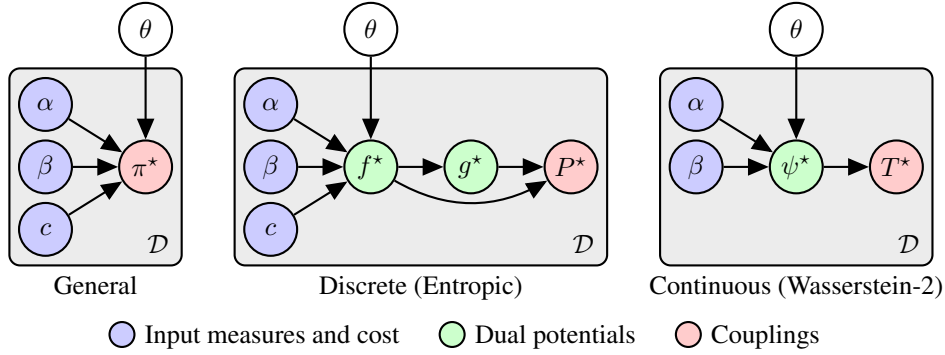
\begin{figure*}[t]
  \centering
  \begin{tikzpicture}[every path/.style={thick}]
  \tikzstyle{obs} = [latent,fill=blue!20];
  \tikzset{plate caption/.append style={below right=-.3cm and -.35cm of #1.south east}};
  \node[obs] (alpha) {$\alpha$};
  \node[obs,below=.1cm of alpha] (beta) {$\beta$};
  \node[obs,below=.1cm of beta] (c) {$c$};
  \node[obs,right=.6cm of beta,fill=red!20] (pi) {$\pi^\star$};
  \node[latent,above=1.1cm of pi] (theta) {$\theta$};
  \scoped[on background layer]{
    \plate [inner sep=.1cm,fill=gray!15] {plate} {(alpha) (beta) (c) (pi)} {};
  };
  \edge {alpha,beta,c} {pi};
  \edge {theta} {pi};
  \node[below right=-5.5mm and -6mm of plate.south east] {$\gD$};
  \begin{pgfinterruptboundingbox}
    \node[below=0mm of plate.south] {General};
  \end{pgfinterruptboundingbox}
\end{tikzpicture}
\hspace{.5cm}
\begin{tikzpicture}[every path/.style={thick}]
  \tikzstyle{obs} = [latent,fill=blue!20];
  \tikzset{plate caption/.append style={below right=-.3cm and -.35cm of #1.south east}};
  \node[obs] (alpha) {$\alpha$};
  \node[obs,below=.1cm of alpha] (beta) {$\beta$};
  \node[obs,below=.1cm of beta] (c) {$c$};
  \node[obs,right=.6cm of beta,fill=green!20] (f) {$f^\star$};
  \node[obs,right=.6cm of f,fill=green!20] (g) {$g^\star$};
  \node[obs,right=.6cm of g,fill=red!20] (P) {$P^\star$};
  \node[latent,above=1.1cm of f] (theta) {$\theta$};
  \scoped[on background layer]{
    \plate [inner sep=.1cm,fill=gray!15] {plate} {(alpha) (beta) (c) (f) (g) (P)} {};
  };
  \edge {alpha,beta,c} {f};
  \edge {theta} {f};
  \edge {f} {g};
  \edge {g} {P};
  \draw[->] (f) to [out=-30,in=-150] (P);
  \node[below right=-5.5mm and -6mm of plate.south east] {$\gD$};
  \begin{pgfinterruptboundingbox}
    \node[below=0mm of plate.south] {Discrete (Entropic)};
  \end{pgfinterruptboundingbox}
\end{tikzpicture}
\hspace{.5cm}
\begin{tikzpicture}[every path/.style={thick}]
  \tikzstyle{obs} = [latent,fill=blue!20];
  \tikzset{plate caption/.append style={below right=-.3cm and -.35cm of #1.south east}};
  \node[obs] (alpha) {$\alpha$};
  \node[obs,below=.1cm of alpha] (beta) {$\beta$};
  \node[obs,below=.1cm of beta,draw=none,fill=none] (c) {};
  \node[obs,right=.6cm of beta,fill=green!20] (psi) {$\psi^\star$};
  \node[obs,right=.6cm of psi,fill=red!20] (T) {$T^\star$};
  \node[latent,above=1.1cm of psi] (theta) {$\theta$};
  \scoped[on background layer]{
    \plate [inner sep=.1cm,fill=gray!15] {plate} {(alpha) (beta) (c) (psi) (T)} {};
  };
  \node[below right=-5.5mm and -6mm of plate.south east] {$\gD$};
  \edge {alpha,beta} {psi};
  \edge {theta} {psi};
  \edge {psi} {T};
  \begin{pgfinterruptboundingbox}
    \node[below=0mm of plate.south] {Continuous (Wasserstein-2)};
  \end{pgfinterruptboundingbox}
\end{tikzpicture}
\\[6mm]
\circlegend{{blue!20}}{Input measures and cost} \hspace{2mm}
\circlegend{{green!20}}{Dual potentials} \hspace{2mm}
\circlegend{{red!20}}{Couplings}
  \vspace{-4mm}
  \caption{
    Meta OT uses objective-based amortization for optimal transport.
    In the general formulation, the \emph{parameters} $\theta$ capture
    shared structure in the \emph{optimal couplings} $\pi^\star$ between
    multiple input measures and costs over some \emph{distribution} $\gD$.
    In practice, we learn this shared structure over the
    \emph{dual potentials} which map back to the coupling:
    $f^\star$ in discrete settings and $\psi^\star$ in continuous ones.
    \vspace*{-3mm}
  }
  \label{fig:meta-OT}
\end{figure*}

Let $\alpha$ and $\beta$ be continuous measures in Euclidean
space $\gX=\gY=\R^d$ \textcolor{gray}{(with $\alpha$ absolutely continuous
with respect to the Lebesgue measure)} and the ground cost be
the squared Euclidean distance
$c(x,y)\defeq\|x-y\|_2^2$.
Then the minimum of \cref{eq:kantorovich-primal} defines
the square of the \emph{Wasserstein-2} distance:
\begin{align}
  W_2^2(\alpha,\beta) \defeq &\min_{\pi\in\gU(\alpha, \beta)}
    \int_{\gX\times \gY} \|x-y\|_2^2{\rm d}\pi(x,y)
     \\
    =&\min_{T}\int_\gX \|x-T(x)\|_2^2{\rm d}\alpha(x),
  \label{eq:W2}
\end{align}
where $T$ is a \emph{transport map} pushing
$\alpha$ to $\beta$, \ie $T_{\#}\alpha=\beta$
with the \emph{pushforward operator}
defined by $T_{\#}\alpha(B) \defeq \alpha(T^{-1}(B))$
for any measurable set $B$.

\textbf{Convex dual potentials.}
The primal in \cref{eq:W2} is difficult to solve
due to the constraints and many computational methods
\citep{makkuva2020optimal,taghvaei2019wasserstein,korotin2019wasserstein,korotin2021continuous,korotin2022neural,amos2022amortizing}
solve the dual
\begin{equation}
  \psi^\star(\ \cdot\ ; \alpha, \beta)\in \argmin_{\psi\in{\rm convex}}\int_\gX \psi(x){\rm d}\alpha(x) + \int_\gY \overline\psi(y){\rm d}\beta(y),
  \label{eq:W2-dual}
\end{equation}
where $\psi$ is a convex function referred to as a \emph{potential},
and $\overline\psi(y)\defeq \max_{x\in\gX} \langle x, y\rangle - \psi(x)$
is the \emph{Legendre-Fenchel transform} or \emph{convex conjugate}
of $\psi$ \citep{fenchel1949conjugate,rockafellar2015convex}.
The potential may be approximated with an
input-convex neural network (ICNN) \citep{amos2017input}.

\textbf{Recovering the primal solution from the dual.}
Given an optimal dual $\psi^\star$ for \cref{eq:W2-dual},
\citet{brenier1991polar} shows that an optimal map
$T^\star$ for \cref{eq:W2} can be obtained with
\begin{equation}
  T^\star(x) = \nabla_x \psi^\star(x).
  \label{eq:W2-dual-to-primal}
\end{equation}

\textbf{Wasserstein-2 Generative Networks (W2GNs).}
\citet{korotin2019wasserstein} model $\psi_\varphi$ and
$\overline{\psi_\varphi}$ in \cref{eq:W2-dual} with two separate
ICNNs parameterized by $\varphi$.
The separate model for $\overline{\psi_\varphi}$ is useful because
the conjugate operation in \cref{eq:W2-dual} becomes
computationally expensive.
They optimize the loss
\newcommand{\detachedphi}{\text{\sout{\ensuremath{\varphi}}}}
\begin{equation}
  \begin{aligned}
  \hspace{-1mm}
  \gL(\varphi) &\defeq
    \underbrace{\E_{x\sim \alpha} \left[ \psi_\varphi(x)\right] +
     \E_{y\sim \beta} \left[ \langle \nabla\overline{\psi_\detachedphi}(y), y\rangle -
\psi_\varphi(\nabla\overline{\psi_\detachedphi}(y)) \right]}_{\text{Cyclic monotone correlations (dual objective)}} \\
 &\hspace{4mm} + \gamma \underbrace{
    \E_{y\sim \beta} \|\nabla\psi_\varphi\circ \nabla\overline{\psi_\varphi}(y)-y\|_2^2,
   }_{\text{Cycle-consistency regularizer}} \\[-16mm]
  \label{eq:w2gn-loss}
  \end{aligned}
\end{equation} \\[-1mm]

where $\detachedphi$ is a detached copy of the parameters
and $\gamma$ is a hyper-parameter.
The first term are the
\emph{cyclic monotone correlations}
\citep{chartrand2009gradient,taghvaei2019wasserstein},
that optimize the dual objective in \cref{eq:W2-dual},
and the second term provides \emph{cycle consistency}
\citep{zhu2017unpaired} to estimate the conjugate $\overline \psi$.
\Cref{alg:w2gn} shows how $\gL$ is typically optimized
using samples from the measures, which we use to
fine-tune Meta OT predictions.

\subsection{Amortized optimization and learning to optimize}
\label{sec:prelim:amor}
Our paper is an application of amortized optimization methods
that predict the solutions of optimization problems,
as surveyed in, \eg, \citet{chen2021learning,amos2022tutorial}.
We use the setup from \citet{amos2022tutorial},
which considers unconstrained continuous optimization problems
\begin{equation}
  z^\star(\phi) \in \argmin_z J(z; \phi),
  \label{eq:amor-star}
\end{equation}
where $J$ is the objective, $z\in\gZ$ is the \emph{domain}, and
$\phi\in\Phi$ is some \emph{context} or \emph{parameterization}.
In other words, the context conditions the objective but
is not optimized over.
Given a \emph{distribution over contexts} $\gP(\phi)$,
we learn a model $\hat z_\theta$ parameterized by $\theta$
to approximate \cref{eq:amor-star}, \ie $\hat z_\theta(\phi)\approx z^\star(\phi)$.
$J$ will be differentiable, so we optimize
the parameters using \emph{objective-based learning} with
\begin{equation}
  \min_\theta \E_{\phi\sim \gP(\phi)} J(\hat z_\theta(\phi); \phi),
  \label{eq:amor-obj-opt}
\end{equation}
which does \emph{not} require ground-truth solutions $z^\star$
and can be optimized with a gradient-based solver.

\section{Meta Optimal Transport}
\label{sec:meta}

We refer to Meta Optimal Transport as the setting when
amortized optimization (\cref{sec:prelim:amor}) is used
for predicting solutions to optimal transport problems
such as \cref{eq:kantorovich-primal}.
We refer to the distribution over the OT problems (measures and costs)
as the \emph{meta-distribution} and denote it as
$\gD(\alpha,\beta,c)$, which we call \emph{meta}
to distinguish it from the measures $\alpha,\beta$.
For example, \cref{sec:exp:mnist,sec:exp:world} considers meta-distributions
over the weights of the atoms, \ie $(a,b)\sim\gD$, where
$\gD$ is a distribution over $\Delta_{m-1} \times \Delta_{n-1}$.
While a model could directly predict the primal
solution to \cref{eq:kantorovich-primal}, \ie
$P_\theta(\alpha, \beta, c)\approx P^\star(\alpha, \beta, c)$
for $(\alpha, \beta, c)\sim \gD$,
this is difficult due to the coupling constraints.
We instead opt to predict the dual variables.
\Cref{fig:meta-OT} illustrates Meta OT in discrete
and continuous settings.

\subsection{Meta OT between discrete measures}
We build on standard methods for entropic OT
reviewed in \cref{sec:prelim:discrete} between discrete measures
$\alpha \defeq \sum_{i=1}^m a_i \delta_{x_i}$ and $\beta \defeq \sum_{i=1}^n b_i \delta_{x_i}$
with $a \in \Delta_{m-1}$ and $b\in \Delta_{n-1}$
coupled using a cost $c$.
In the Meta OT setting, the measures and cost are the contexts for
amortization and sampled from a
\emph{meta-distribution}, \ie $(\alpha,\beta,c)\sim\gD(\alpha,\beta,c)$.
For example, \cref{sec:exp:mnist,sec:exp:world} considers meta-distributions
over the weights of the atoms, \ie $(a,b)\sim\gD$, where
$\gD$ is a distribution over $\Delta_{m-1} \times \Delta_{n-1}$.

\textbf{Amortization objective.}
We will seek to predict the \emph{optimal} potential.
At optimality, the pair of potentials are related to each
other via \cref{eq:f-to-g}, \ie
$g(f; \alpha, \beta, c) \defeq \epsilon \log b - \epsilon\log\left(K^\top \exp\{f/\epsilon\}\right)$
where $K\in\R^{m\times n}$ is the \emph{Gibbs kernel}
from \cref{eq:entropic-dual}.
Hence, it is sufficient to predict one of the
potentials, \eg $f$, and recover the other.
We thus re-formulate \cref{eq:entropic-dual}
to just optimize over $f$ with
\begin{equation}
  f^\star(\alpha, \beta, c, \epsilon)\in\ \argmin_{f\in\R^n}\ J(f; \alpha, \beta, c),
  \label{eq:entropic-dual-f}
\end{equation}
where $-J(f; \alpha, \beta, c)\defeq \langle f, a\rangle + \langle g, b \rangle
    - \epsilon\left\langle \exp\{f/\epsilon\}, K\exp\{g/\epsilon\}\right\rangle$
is the (negated) dual objective.
Even though most solvers optimize over $f$ and $g$ jointly
as in \cref{eq:entropic-dual-f}, amortizing over these would likely need to
have a higher capacity than a model just predicting $f$ and
learn how $f$ and $g$ are connected through \cref{eq:f-to-g} while
in \cref{eq:entropic-dual-f} we explicitly provide this knowledge.

\textbf{Amortization model.}
We predict the solution to \cref{eq:entropic-dual-f} with
$\hat f_\theta(\alpha, \beta, c)$ parameterized by $\theta$, resulting in
a computationally efficient approximation $\hat f_\theta \approx f^\star$.
Here we use the notation $\hat f_\theta(\alpha,\beta,c)$ to mean that
the model $\hat f_\theta$ depends on \emph{representations} of
the input measures and cost.
In our settings, we define $\hat f_\theta$ as a fully-connected MLP
mapping from the atoms of the measures to the duals.

\textbf{Amortization loss.}
Applying objective-based amortization from \cref{eq:amor-obj-opt} to
the dual in \cref{eq:entropic-dual-f} completes the
learning setup.
The model should optimize the expected dual value:
\begin{equation}
  \min_\theta \E_{(\alpha,\beta,c)\sim\gD} J(\hat f_\theta(\alpha, \beta, c); \alpha, \beta, c),
  \label{eq:amor-entropic-loss}
\end{equation}
which is appealing as it does not require ground-truth solutions $f^\star$.
The ground-truth solutions may be expensive to obtain, but if they
are available, a regression term can also be added
\citep{amos2022tutorial}.
\Cref{alg:meta-ot-training} shows a basic training loop for
\cref{eq:amor-entropic-loss} using a gradient-based
optimizer such as Adam \citep{kingma2014adam}.

\textbf{Sinkhorn fine-tuning.}
The dual prediction made by $\hat f_\theta$ with an associated $\hat g$
can be used to initialize a standard Sinkhorn solver.
This allows for the predicted solution to be refined
to an optimality threshold.

\textbf{On accelerated solvers.}
While we have considered fine-tuning the Meta OT prediction
with a log-Sinkhorn solver,
Meta OT can also be combined with accelerated variants of
entropic OT solvers such as
\citet{thibault2017overrelaxed,altschuler2017near,alaya2019screening,lin2019acceleration}
that otherwise solve every problem from scratch.

\begin{algorithm}[t]
    \caption{Training Meta OT}
    \begin{algorithmic}
    \footnotesize
    \State Initialize amortization model with $\theta_0$
    \For{iteration $i$}
    \State Sample $(\alpha, \beta, c)\sim\gD$
    \State Predict duals $\hat f_\theta$ or $\hat \varphi_\theta$ on the sample
    \State Estimate the loss in \cref{eq:amor-entropic-loss} or \cref{eq:amor-w2gn-loss}
    \State Update $\theta_{i+1}$ with a gradient step
    \EndFor
    \end{algorithmic}
    \label{alg:meta-ot-training}
\end{algorithm}

\begin{figure*}[t]
  \centering
  \begin{minipage}[t]{0.49\linewidth}
    \centering
    {\large Sinkhorn \color{gray}{(converged, ground-truth)}} \\
    \begin{tikzpicture}[every path/.style={thick}]
      \node[align=left,anchor=south west] {\includegraphics[height=23px]{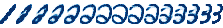}};
      \node at (2.6mm,0) (alpha) {\large$\alpha_0$};
      \node at (36mm,-2mm) {\large$\alpha_1$};
      \draw[line width=.5mm] (36mm,0mm) -- (36mm,2mm);
      \node at (64.5mm,0) (beta) {\large$\alpha_2$};
      \edge[<->] {alpha} {beta};
    \end{tikzpicture}
  \end{minipage}
  \hfill
  \begin{minipage}[t]{0.49\linewidth}
    \centering
    {\large Meta OT \color{gray}{(initial prediction)}} \\
    \begin{tikzpicture}[every path/.style={thick}]
      \node[align=left,anchor=south west] {\includegraphics[height=23px]{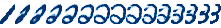}};
      \node at (2.6mm,0) (alpha) {\large$\alpha_0$};
      \node at (35mm,-2mm) {\large$\alpha_1$};
      \draw[line width=.5mm] (35mm,0mm) -- (35mm,2mm);
      \node at (64mm,0) (beta) {\large$\alpha_2$};
      \edge[<->] {alpha} {beta};
    \end{tikzpicture}
  \end{minipage}
  \caption{
    Interpolations between MNIST test digits using
    couplings obtained from
    (left) solving the problem with Sinkhorn, and
    (right) Meta OT model's initial prediction, which
    is $\bf{\approx}100$ times computationally cheaper and produces a nearly identical coupling.
  }
  \label{fig:mnist-test-vis}
\end{figure*}

\begin{table}[t]
  \newcommand{\pair}[2]{$#1$ {\color{gray}\footnotesize $\pm #2$}}
    \centering
    \caption{Sinkhorn runtime (seconds) to reach a marginal error
    of $10^{-2}$.
    Meta OT's initial prediction takes $\approx 5\cdot10^{-5}$ seconds.
    We report the mean {\color{gray}and std} across 10 test instances.
    }
    \resizebox{\linewidth}{!}{
    \begin{tabular}{r|ll} \toprule
    Initialization & MNIST & Spherical \\\midrule
    Zeros ($t_{\rm zeros}$) & \pair{4.5\cdot10^{-3}}{1.5\cdot10^{-3}} & \pair{0.88}{0.13} \\
    Gaussian & \pair{4.1\cdot10^{-3}}{1.2\cdot10^{-3}} & \pair{0.56}{9.9\cdot10^{-2}} \\
    Meta OT ($t_{\rm Meta}$) & \cellhi \pair{2.3\cdot10^{-3}}{9.2\cdot10^{-6}} & \cellhi \pair{7.8\cdot10^{-2}}{3.4\cdot10^{-2}} \\ \midrule
    Improvement ($t_{\rm zeros}/t_{\rm Meta}$) & 1.96 & 11.3 \\
    \bottomrule
    \end{tabular}}
    \label{tab:discrete-runtime}
\end{table}

\subsection{Meta OT between continuous measures}
\label{sec:meta-ot:icnn}

We take an analogous approach to predicting the Wasserstein-2
map between continuous measures for Wasserstein-2 as reviewed in
\cref{sec:prelim:continuous}.
Here the measures $\alpha,\beta$ are supported in continuous space
$\gX=\gY=\R^d$ and we focus on computing Wasserstein-2 couplings
from instances sampled from a \emph{meta-distribution}
$(\alpha,\beta)\sim\gD(\alpha,\beta)$.
The cost $c$ is not included in $\gD$ as it remains fixed to the
squared Euclidean cost everywhere here.

One challenge here is that the optimal dual potential
$\psi^\star(\ \cdot\ ; \alpha, \beta)$
in \cref{eq:W2-dual} is a convex function and not simply
a finite-dimensional real vector.
The dual potentials in this setting are approximated by, \eg, an ICNN.
We thus propose a \emph{Meta ICNN} that predicts the \emph{parameters}
$\varphi$ of an ICNN $\psi_\varphi$ that approximates the optimal
dual potentials, which can be seen as a hypernetwork
\citep{stanley2009hypercube,ha2016hypernetworks}.
The dual prediction made by $\hat \varphi_\theta$
can easily be input as the initial value to a standard W2GN solver.
\Cref{app:other-W2-models} discusses other modeling
choices we considered: we tried models based on
MAML \citep{pmlr-v70-finn17a} and neural processes
\citep{garnelo2018neural,garnelo2018conditional}.

\begin{figure*}[t]
  \centering
  \includegraphics[width=0.42\textwidth]{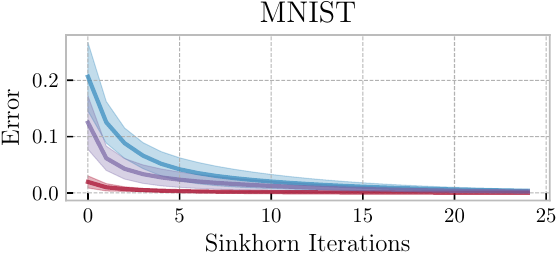}
  \hspace{2mm}
  \includegraphics[width=0.42\textwidth]{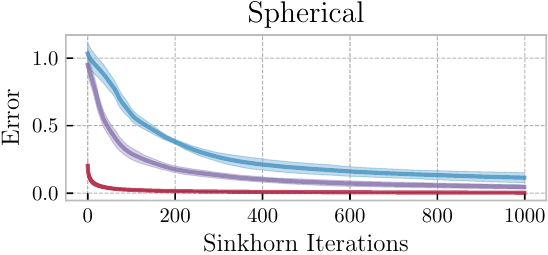} \\[2mm]
  Initialization (\cblock{52}{138}{189} Zeros \hspace{2mm}
  \cblock{122}{104}{166} Gaussian \citep{thornton2022rethinking} \hspace{2mm}
  \cblock{166}{6}{40} Meta OT)
  \vspace{-3mm}
  \caption{
    Meta OT successfully predicts warm-start initializations
    that significantly improve the convergence of Sinkhorn iterations on test data.
    The error is the marginal error defined in \cref{eq:sinkhorn-err}.
  }
  \label{fig:discrete-errs}
\end{figure*}

\textbf{Amortization objective.}
We build on the W2GN formulation \citep{korotin2019wasserstein}
and seek parameters $\varphi^\star$ optimizing the
dual ICNN potentials $\psi_\varphi$ and $\overline{\psi_\varphi}$
with $\gL(\varphi; \alpha, \beta)$ from \cref{eq:w2gn-loss}.
We chose W2GN due to the stability, but could also easily
use other losses optimizing ICNN potentials.

\textbf{Amortization model: the Meta ICNN.}
We predict the solution to \cref{eq:w2gn-loss} with
$\hat \varphi_\theta(\alpha, \beta)$ parameterized by $\theta$,
resulting in a computationally efficient approximation to
the optimum $\hat \varphi_\theta \approx \varphi^\star$.
\Cref{fig:meta-icnn} instantiates a convolutional Meta ICNN
model using a ResNet-18 \citep{he2016identity} architecture for
coupling image-based measures.
We again emphasize that $\alpha,\beta$ used with the model
here are \emph{representations} of measures, which in our cases are simply images.

\textbf{Amortization loss.}
Applying objective-based amortization from \cref{eq:amor-obj-opt} to
the W2GN loss in \cref{eq:w2gn-loss} completes our learning setup.
We optimize the loss
\begin{equation}
  \min_\theta \E_{(\alpha,\beta)\sim\gD}
    \gL(\hat \varphi_\theta(\alpha, \beta); \alpha, \beta).
  \label{eq:amor-w2gn-loss}
\end{equation}
As in the discrete setting, this loss does not require
ground-truth solutions $\varphi^\star$ and we find
the solution with Adam.

\section{Experiments}\label{sec:experiments}
We demonstrate how Meta OT models improve the convergence
of the state-of-the-art solvers in settings where
solving multiple OT problems naturally arises.
We implemented our code in JAX \citep{jax2018github} as an
extension to the the Optimal Transport Tools (OTT)
package \citep{cuturi2022optimal}.
\Cref{app:exp-details} covers further experimental
and implementation details, and shows that all of our
experiments take a few hours to
run on our single Quadro GP100 GPU.
The source code to reproduce all of our experiments is available at
\url{http://github.com/facebookresearch/meta-ot}.

\begin{figure*}[t]
  \centering
  \begin{minipage}{0.45\linewidth}
    \centering
    {\large Sinkhorn \color{gray}{(converged, ground-truth)}} \\
    \includegraphics[width=\textwidth]{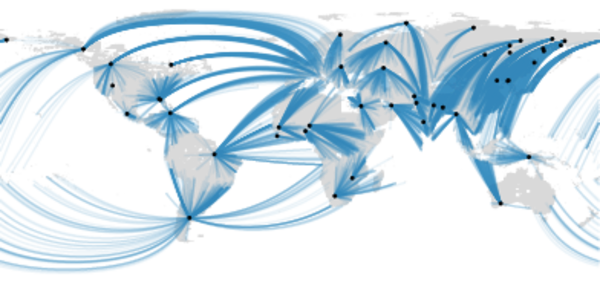}
    \includegraphics[width=\textwidth]{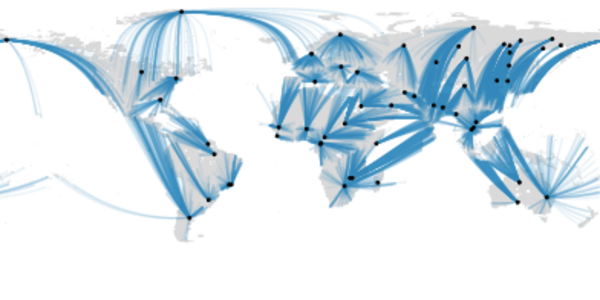}
  \end{minipage}
  \hspace{3mm}
  \begin{minipage}{0.45\linewidth}
    \centering
    {\large Meta OT \color{gray}{(initial prediction)}} \\
    \includegraphics[width=\textwidth]{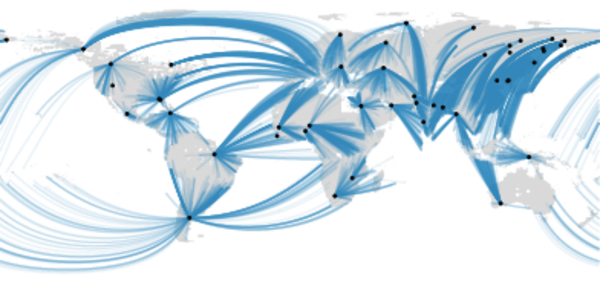}
    \includegraphics[width=\textwidth]{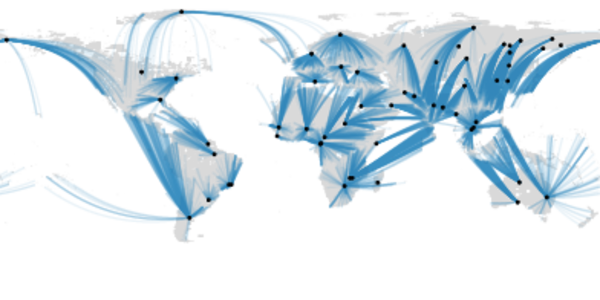}
  \end{minipage}
  \caption{Test set coupling predictions of the spherical transport problem.
    Meta OT's initial prediction is $\bf{\approx}37500$ times faster
    than solving Sinkhorn to optimality.
    Supply locations are shown as black dots and the blue lines
    show the spherical transport maps $T$ going to demand locations
    at the end.
    The sphere is visualized with the Mercator projection.
  }
  \label{fig:world-errs}
\end{figure*}

\subsection{Discrete OT}
\subsubsection{Grayscale image transport}
\label{sec:exp:mnist}
Images provide a natural setting for Meta OT where the
distribution over images provide the meta-distribution
$\gD$ over OT problems.
Given a pair of images $\alpha_0$ and
$\alpha_1$, each grayscale image is cast as a discrete measure
in 2-dimensional space where the intensities define the probabilities of the atoms.
The goal is to compute the optimal transport interpolation between the
two measures as in, \eg, \citet[\S7]{peyre2019computational}.
Formally, this means computing the optimal coupling $P^\star$
by solving the entropic optimal transport problem between $\alpha_0$ and
$\alpha_1$ and computing the interpolates as
$\alpha_t = (t\proj_y+(1-t)\proj_x) _{\#}P^\star$, for $t\in [0,1]$,
where $\proj_x(x,y)\defeq x$ and $\proj_y(x,y)\defeq y$.
We selected $\epsilon=10^{-2}$ as \cref{app:mnist-eps} shows that
it gives interpolations that are not too blurry or sharp.

Our Meta OT model $\hat f_\theta$ (\cref{sec:meta})
is an MLP that predicts the transport map between pairs of MNIST digits.
We train on every pair from the standard training dataset.
\Cref{fig:mnist-test-vis} shows that even without fine-tuning,
Meta OT's predicted Wasserstein interpolations between
the measures are close to the ground-truth interpolations
obtained from running the Sinkhorn algorithm to convergence.
We then fine-tune Meta OT's prediction with Sinkhorn.
\Cref{fig:discrete-errs} shows that the near-optimal predictions
can be quickly refined in fewer iterations than running
Sinkhorn with the default initialization,
and \cref{tab:discrete-runtime} shows the runtime required
to reach an error threshold of $10^{-2}$, showing that the Meta OT
initialization help solve the problems faster by an order of magnitude.
We compare our learned initialization to the standard zero initialization,
as well as the Gaussian initialization proposed in \citet{thornton2022rethinking},
which takes a continuous Gaussian approximation of the measures and initializes
the potentials to be the known coupling between the Gaussians.
This Gaussian initialization assumes the squared Euclidean cost,
which is not the case in our spherical transport problem,
but we find it is still helpful over the zero initialization.

\textbf{Out-of-distribution generalization}
We now test the ability of Meta OT to
predict potentials for out-of-distribution input data.
We consider the pairwise training and evaluation on the following datasets:
1) MNIST; 2) USPS \citep{uspsdataset} (upscaled to have the same size as the MNIST);
3) Google Doodles dataset\footnote{\url{https://quickdraw.withgoogle.com/data}}
with classes Crab, Cat and Faces;
4) sparsified random uniform data in [0,1] where
sparsity (zeroing values below 0.95) is used to mimic the sparse
signal in black-and-white images.
For each pair, eg, MNIST-USPS, we train on one dataset and use
the other to predict the potentials.
The comparison is done using the same metric as before, \ie,
the deviation from the marginal constraints
defined in \cref{eq:sinkhorn-err}.
\Cref{fig:cross_domain} shows how well the learned models
are capable of transferring to new domains.

\subsubsection{Supply-demand transportation on spherical data}
\label{sec:exp:world}

We next set up a synthetic transport problem between supply and demand
locations where the supply and demands may change locations or quantities
frequently, creating another Meta OT setting to be able to rapidly solve
the new instances.
We specifically consider measures living on the 2-sphere defined by
$\gS_2\defeq\{x\in\R^3: \|x\|=1\}$, \ie $\gX=\gY=\gS_2$,
with the transport cost given by the spherical distance
$c(x,y)=\arccos(\langle x, y \rangle)$.
We then randomly sample supply locations uniformly from Earth's
landmass and demand locations from Earth's population density to
induce a class of transport problems on the sphere
obtained from the CC-licensed dataset from \citet{doxsey2015taking}.
\Cref{fig:world-errs} shows that the predicted transport maps
on test instances are close to the optimal maps obtained from
Sinkhorn to convergence.
Similar to the MNIST setting, \cref{fig:discrete-errs,tab:discrete-runtime}
show improved convergence and runtime.

\begin{figure*}
\centering
\hspace*{-3mm}
    \includegraphics[height=1.5in]{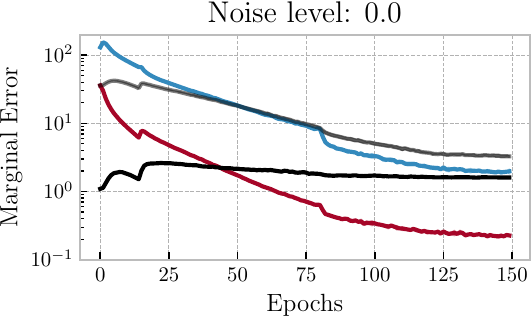}\hspace{-1.3mm}
    \includegraphics[height=1.5in]{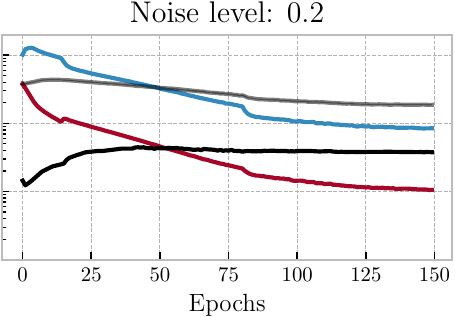}\hspace{-1.3mm}
    \includegraphics[height=1.5in]{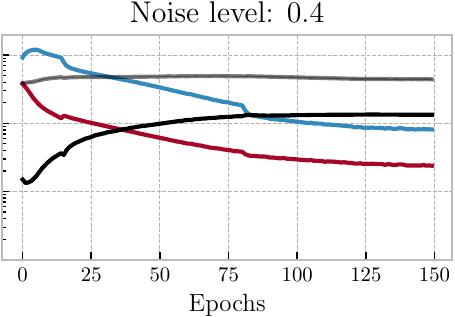} \\
    Zero initialization (\cblock{90}{90}{90} +5 iterations \hspace{2mm} \cblock{0}{0}{0} +20 iterations) \hspace{3mm}
    Meta OT (\cblock{52}{138}{189} Prediction \hspace{2mm} \cblock{166}{6}{40} +5 iterations)
    \label{fig:war_marginal_errors}
    \caption{Marginal errors throughout WAR training for CIFAR-100 classification.
      The bumps correspond to when the loss and learning rate are updated during
      training as described in \citet{fatras2021wasserstein}}
\end{figure*}

\subsubsection{Wasserstein adversarial regularization}
\label{sec:exp:war}
\begin{table*}[t!]
    \newcommand{\pair}[2]{$#1$ {\color{gray}\footnotesize $\pm #2$}}
    \centering
    \resizebox{0.65\linewidth}{!}{
    \begin{tabular}{c |c | c c c}
    & & \multicolumn{3}{c}{Methods} \\

          Dataset & Noise level & WAR (5 iter.) & WAR (20 iter.) & Meta OT + 5 iter. \\
          \midrule
         \multirow{3}{*}{Fashion MNIST}     & 0\% & \pair{94.75}{0.05} & \pair{94.16}{0.02} & \pair{94.70}{0.05}\\ 
                                            & 20\% & \pair{93.00}{0.15} & \pair{93.41}{0.08} & \pair{93.53}{0.01} \\
                                            & 40\% & \pair{88.67}{0.10} & \pair{88.08}{0.10} & \pair{89.08}{0.61}\\
                                            \midrule
         \multirow{3}{*}{Cifar-10} & 0\% & \pair{91.96}{0.13} & \pair{91.76}{0.25} & \pair{91.98}{0.16} \\ 
                                            & 20\% & \pair{88.80}{0.11} & \pair{90.59}{0.05} & \pair{90.13}{0.21} \\
                                            & 40\% & \pair{81.09}{0.06} & \pair{87.07}{0.08} & \pair{83.57}{1.13} \\
                                            \midrule
    \multirow{3}{*}{Cifar-100} & 0\% & \pair{70.93}{0.23} & \pair{69.79}{0.34} & \pair{70.25}{0.04} \\ 
                                            & 20\% & \pair{66.23}{0.29} & \pair{66.18}{0.18} & \pair{66.59}{0.38} \\
                                            & 40\% & \pair{52.69}{0.12} & \pair{61.63}{0.33} & \pair{61.13}{0.17} \\
    \end{tabular}
    }
    \caption{Comparison of the original WAR implementation with WAR implementation using only 5 Sinkhorn iterations and our Meta OT model with 5 Sinkhorn iterations on top of initial predictions. We report the mean {\color{gray}and std} across 3 random seeds.}
    \label{tab:war_comparison}
\end{table*}
Wasserstein losses has recently attracted a considerable attention in
the field of multi-label
\citep{Wasserstein15,kdd_wass_loss,rot,Toyokuni2021ComputationallyEW}
and multi-class classification
\citep{liu2020importance_wass_loss,seg_wass_loss,pose_est_wass_loss,obj_det_wass_loss,fatras2021wasserstein}
as they both require finding an informative way of comparing discrete
distributions given by the true labeling of the data points and those
predicted by the classification model. In this experiment, we aim to
show that meta OT model can be learned alongside the training of the
multi-class classification model and used to make predictions for the
Wasserstein loss term appearing in the objective function of the
latter. For this, we consider as an example the setup of
\citet{fatras2021wasserstein} where the authors define an adversarial
loss term (called WAR) aiming at limiting the effect of label noise on
the generalization capacity of deep vision neural networks. In
particular, given a neural network $p_\theta$ predicting a vector of
class memberships in $\R^c$, the regularization term is
\begin{equation}
\begin{aligned}
    R_{\text{WAR}} (x_i) = W_C^\epsilon(p_\theta(x_i + r_i^a), p_\theta(x_i))\\
    r_i^a = \argmax_{r_i, ||r_i||\leq \varepsilon} W_C^\epsilon(p_\theta(x_i + r_i), p_\theta(x_i)).
\end{aligned}
\label{eq:war}
\end{equation}
where $W_C^\epsilon$ is the Wasserstein distance with entropic
regularization introduced in \cref{eq:entropic-primal} with a cost
matrix $C \in R^{c\times c}$. Learning $p_\theta$ is done by optimizing
the cross entropy loss together with $R_{\text{WAR}} (x_i)$ using
stochastic optimization.
This means that OT problems in \cref{eq:war}
are solved repeatedly, for every batch in the input dataset and during
multiple epochs thus making the meta OT warm-starts particularly
computationally attractive in this context. For this task, we optimize
a meta OT model defined as a MLP with 3 hidden layers over the same
data alongside the main optimization procedure. We use meta OT model
to predict the solutions to both OT problems in \cref{eq:war} and use
only 25\% of iterations in the Sinkhorn loop to compute
$W_C^\epsilon$. As in \citet{fatras2021wasserstein}, we evaluate the
efficiency of such learning strategy on three computer vision
datasets, namely: Fashion MNIST, Cifar-10 and Cifar-100. For each of
them, we consider the clean version of the data (0\% noise), and two
variations with 20\% and 40\% of noise in labels. The authors of
\citet{fatras2021wasserstein} experiment with two cost matrices: one is
defined based on the distances between the class centroids of 30000
samples from the original dataset when embedded with ResNet18; second
one is defined as the Euclidean distance between the word2vec
embeddings of the classes of the original dataset. To show the
versatility of our approach with respect to different geometries, we
use the first cost matrix for Fashion MNIST dataset, and the second
one for Cifar-10 and Cifar-100 datasets.

We evaluate meta OT for this task based on three criteria. First, we
want to make sure that reducing the number of iterations in the
Sinkhorn loop is not detrimental for the overall performance of the
learned classification model. These results are presented in
\cref{tab:war_comparison}, where we can see that meta OT leads to the
same performance as the original WAR model while doing only 5
iterations of Sinkhorn on top of the initial predictions. Second, we
show in \cref{fig:war_marginal_errors} that our meta OT model
predicts warm-start initializations that have a low marginal error so
that even its initial predictions are become at least as qualitative
as the solution obtained using 20 iterations of the Sinkhorn
algorithm. Finally, we show in \cref{tab:war-runtimes}, that
training a meta OT model alongside the main model doesn't introduce
any additional overhead in terms of computational time. In this table,
we compare the average runtime of each of the considered baselines and
account for the time needed to make a backward pass for the meta OT
model. This result is important as once a meta OT model is trained, it
can be further used to make predictions without any finetuning for
other training runs with different hyperparameters leading to an
important reduction in terms of computational time.

\begin{table}[t]
  \newcommand{\pair}[2]{$#1$ {\color{gray}\footnotesize $\pm #2$}}
  \centering
  \caption{Runtime (s) per epoch of training of the WAR model.}
  \resizebox{\linewidth}{!}{
    \begin{tabular}{r|llll} \toprule
      & Iter & Fashion MNIST & Cifar-10 & Cifar-100 \\\midrule
      Zero Init & 5 & \pair{11.51}{0.07} & \pair{12.88}{0.02} & \pair{13.13}{0.02} \\
      Meta OT  & 5 & \pair{14.37}{0.04} & \pair{14.07}{0.15} &\pair{14.17}{0.02} \\
      Zero Init & 20 & \pair{17.04}{0.12} & \pair{14.36}{0.02} & \pair{14.37}{0.02} \\
      \bottomrule
    \end{tabular}}
  \label{tab:war-runtimes}
\end{table}

\begin{figure*}[t!]
  \centering
  {\Large $\alpha$ \hspace{1.1in} $\beta$ \hspace{1.1in} $T_\#\alpha$ \hspace{.9in} $T_\#^{-1}\beta$}
  \begin{minipage}{0.8\linewidth}
  \begin{tikzpicture}[node/.style={inner sep=0,outer sep=0}]
    \node[align=left,anchor=north west] {\includegraphics[width=\textwidth]{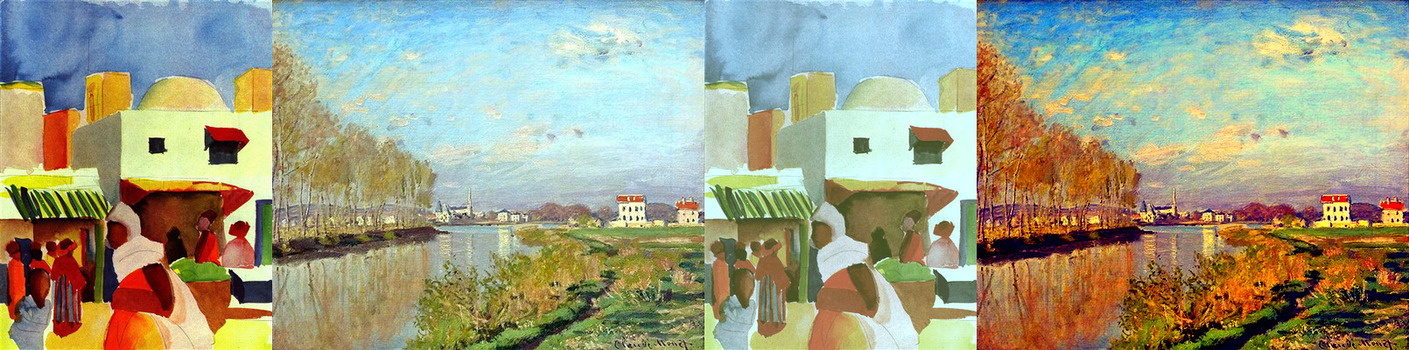}};
    \node[fill=white, opacity=0.75, anchor=north west,inner sep=1mm] at (0,-1mm) {W2GN \color{black!80}{(converged, ground-truth)}};
  \end{tikzpicture} \\[-2.7mm]
  \begin{tikzpicture}[node/.style={inner sep=0,outer sep=0}]
    \node[align=left,anchor=north west] {\includegraphics[width=\textwidth]{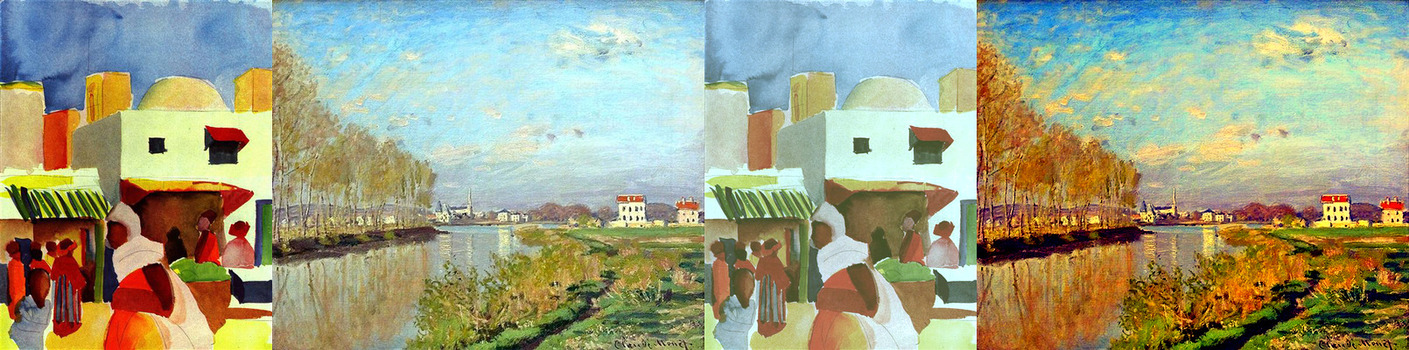}};
    \node[fill=white, opacity=0.75, anchor=north west,inner sep=1mm] at (0,-1mm) {Meta OT \color{black!80}{(Initial prediction)}};
  \end{tikzpicture}
\end{minipage}
  \caption{Color transfers with a Meta ICNN on test pairs of images.
    The objective is to optimally transport the continuous RGB measure
    of the first image $\alpha$ to the second $\beta$,
    producing an invertible transport map $T$.
    Meta OT's prediction is $\bf{\approx}1000$ times faster
    than training W2GN \citep{korotin2019wasserstein} from
    scratch.
    The image generating $\alpha$ is
    \href{https://www.wikiart.org/en/august-macke/market-in-algiers}{Market in Algiers by August Macke (1914)}
    and
    $\beta$ is
    \href{https://www.wikiart.org/en/claude-monet/argenteuil-the-sein}{Argenteuil, The Seine by Claude Monet (1872)},
    obtained from
    \href{https://www.wikiart.org/}{WikiArt}.
  }
  \label{fig:transfer-test-images}
\end{figure*}

\begin{table}[t]
  \newcommand{\pair}[2]{$#1$ {\color{gray}\footnotesize $\pm #2$}}
    \centering
    \caption{Color transfer runtimes and values.
    We report the mean {\color{gray}and std} across 10 test instances.}
  \vspace{1mm}
  \resizebox{\linewidth}{!}{
    \begin{tabular}{r|lll} \toprule
        & Iter & Runtime (s) & Dual Value \\\midrule
Meta OT & None & \pair{3.5\cdot10^{-3}}{2.7\cdot10^{-4}} & \pair{0.90}{6.08\cdot10^{-2}} \\
+ W2GN & 1k & \pair{0.93}{2.27\cdot10^{-2}} & \pair{1.0}{2.57\cdot10^{-3}} \\
& 2k & \pair{1.84}{3.78\cdot10^{-2}} & \pair{1.0}{5.30\cdot10^{-3}} \\ \midrule
W2GN & 1k & \pair{0.90}{1.62\cdot10^{-2}} & \pair{0.96}{2.62\cdot10^{-2}} \\
& 2k & \pair{1.81}{3.05\cdot10^{-2}} & \pair{0.99}{1.14\cdot10^{-2}} \\
        \bottomrule
    \end{tabular}}
    \label{tab:color-runtimes}
\end{table}
\begin{figure}[t]
  \centering
  \includegraphics[width=\linewidth]{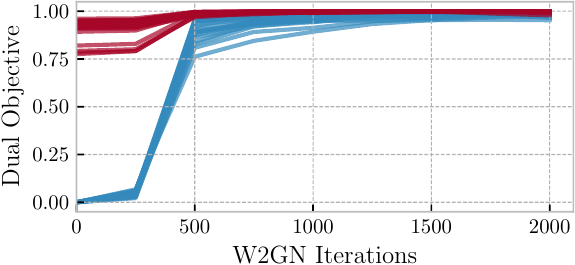} \\
  Initialization (\cblock{52}{138}{189} Standard \hspace{2mm} \cblock{166}{6}{40} Meta OT)
  \vspace*{-3mm}
  \caption{Convergence on color transfer test instances using W2GN.
    Meta ICNNs predicts warm-start initializations
    that significantly improve the (normalized) dual objective values.
  }
  \label{fig:transfer-convergence}
  \vspace*{-3mm}
\end{figure}

\subsection{Continuous OT for color transfer}
The problem of color transfer between two images consists in mapping
the color palette of one image into the other one.
The images are required to have the same number of channels, for
example RGB images.
The continuous formulation that we use from \citet{korotin2019wasserstein},
takes \ie $\gX=\gY=[0,1]^3$ with $c$ being the squared Euclidean distance.
We collected ${\approx}200$ public domain images from
\href{https://www.wikiart.org/}{WikiArt}
and trained a Meta ICNN model from \cref{sec:meta-ot:icnn}
to predict the color transfer maps between every pair of them.
\Cref{fig:transfer-test-images} shows the predictions on test pairs
and \cref{fig:transfer-convergence} shows the convergence in comparison
to the standard W2GN learning.
\Cref{tab:color-runtimes} reports runtimes.
\Cref{app:color} show additional color transfer
\Cref{app:other-W2-models}

\section{Related work}\label{sec:related_work}
\textbf{Efficiently estimating OT maps.}
To compute OT maps with fixed cost between pairs of measures
efficiently, neural OT models
\citep{korotin2019wasserstein,korotin2021neural,mokrov2021large,korotin2021continuous}
leverage ICNNs to estimate maps between continuous high-dimensional
measures given samples from these, and
\citet{litvinenko2021computing,scetbon2021low,forrow2019statistical,sommerfeld2019optimal,scetbon2022linear,muzellec2019subspace,bonet2021subspace}
leverage structural assumptions on coupling and cost matrices to reduce
the computational and memory complexity. In the meta-OT setting, we
consider learning to rapidly compute OT mappings between new pairs
measures. All these works can hence benefit from an
acceleration effect with amortization.

\textbf{Embedding measures where OT distances are discriminative.}
Effort has been invested in learning encodings/projections of measures
through a nested optimization problem, which aims to find
discriminative embeddings of the measures to be compared
\citep{pmlr-v84-genevay18a, deshpande2019max,nguyen2022amortized}.
While these works share an encoder and/or a projection across task
with the aim of leveraging more discriminative alignments (and hence
an OT distance with a metric different from the Euclidean metric), our
work differs in the sense that we find good initializations to solve
the OT problem itself with fixed cost more efficiently across tasks.

\textbf{Optimal transport and amortization.}
\citet{courty2018learning} learn a latent space in which
the Wasserstein distance between the measure's embeddings is
equivalent to the Euclidean distance.
\citet{nguyen2022amortized} amortizes the estimation of
the optimal projection in the max-sliced objective, which differs from our work
where we instead amortize the estimation of the optimal coupling
directly.
\citet{lacombe2021learning} learns to predict
Wasserstein barycenters of pixel images by training a convolutional
networks that, given images as input, outputs their barycenters. Our
work is hence a generalization of this pixel-based work to general
measures -- both discrete and continuous.
One limitation is that the barycenter predictions do not
provide the optimal couplings.
\citet{gracyk2022geonet} learn a neural operator,
\eg from \citet{kovachki2021neural,li2020neural}
to amortize the solution to the PDE from the dynamic OT formulation.
\citet{bunne2022supervised} predict the solutions to continuous
neural OT problems.

\section{Conclusions}
\label{sec:con}
We have presented foundations for modeling and learning to
solve OT problems with Meta OT by using amortized optimization
to predict optimal transport plans.
This works best in applications that require solving
multiple OT problems with shared structure.
We instantiated it to speed up entropic regularized optimal
transport and unregularized optimal transport with squared
cost by multiple orders of magnitude.
We envision extensions of the work in:
1) \textbf{Continuous settings}.
  Learning solutions continuous OT problems is a budding topic
  in the community:
  \citet{gracyk2022geonet} amortize solutions to dynamic OT
  problems between continuous measures, and
  \citet{bunne2022supervised} uses a partially
  input-convex neural network (PICNN) from \citet{amos2017input}
  to predict continuous OT solutions from contextual information.
  Related to these, \cref{sec:appendix:continuous} presents
  a more general extension of Meta OT and provides a demonstration
  on transferring color palettes, which is shown in
  \cref{fig:transfer-test-images}.
  Future directions for amortizing continuous OT problems
  include exploring modeling (PICNN vs.~a hypernetwork),
  loss, and fine-tuning choices.
2) \textbf{Meta OT models}.
While we mostly consider models based on hypernetworks,
other meta-learning paradigms can be connected in.
In the discrete setting, we only considered settings where
the cost remains fixed, but the Meta OT model can also be conditioned
on the cost by considering the entire cost matrix as an input
(which may be too large for most models to handle), or considering
a lower-dimensional parameterization of the cost that changes between
the Meta OT problem instances.
Another modeling dimension is the ability to capture variable-length
input measures. Design decisions for this can be inspired from
by VeLO \citep{metz2022velo}, which learns a generic
optimizer for large-scale machine learning models that
can predict updates to models with 500M parameters.
3) \textbf{OT algorithms}. While we instantiated
models on top of log-Sinkhorn, Meta OT
could be built on top of other methods, and
4) \textbf{OT applications} that are computationally
expensive and repeatedly solved, \eg in multi-marginal
and barycentric settings, or for Gromov-Wasserstein
distances between metric-measure spaces.

\textbf{Limitations.}
While we have illustrated successful applications of Meta OT,
it is also important to understand the limitations that
also arise in more general amortization settings:
1) \textbf{Meta OT does not make previously intractable
problems tractable.}
All of the baseline OT solvers we consider solve
our problems within milliseconds or seconds.
2) \textbf{Out-of-distribution generalization.}
Meta OT may not generate good predictions on instances
that are not close to the training OT problems from the
meta-distribution $\gD$ over the measures and cost.
If the model makes a bad prediction,
one fallback option is to re-solve the instance from scratch.
\looseness=-1

\subsection*{Acknowledgments}
We would like to thank
Eugene Vinitsky,
Mark Tygert,
Mathieu Blondel,
Maximilian Nickel, and
Muhammad Izzatullah
for insightful comments and discussions.
The core set of tools in Python
\citep{van1995python,oliphant2007python}
enabled this work, including
Hydra \citep{Yadan2019Hydra},
JAX \citep{jax2018github},
Matplotlib \citep{hunter2007matplotlib},
numpy \citep{oliphant2006guide,van2011numpy},
Optimal Transport Tools \citep{cuturi2022optimal},
and pandas \citep{mckinney2012python}.

\newpage
\bibliographystyle{plainnat}
\bibliography{refs}

\newpage
\appendix
\onecolumn

\section{Selecting $\epsilon$ for MNIST}
\label{app:mnist-eps}
\begin{figure}[h]
  \centering
  \includegraphics[width=0.8\textwidth]{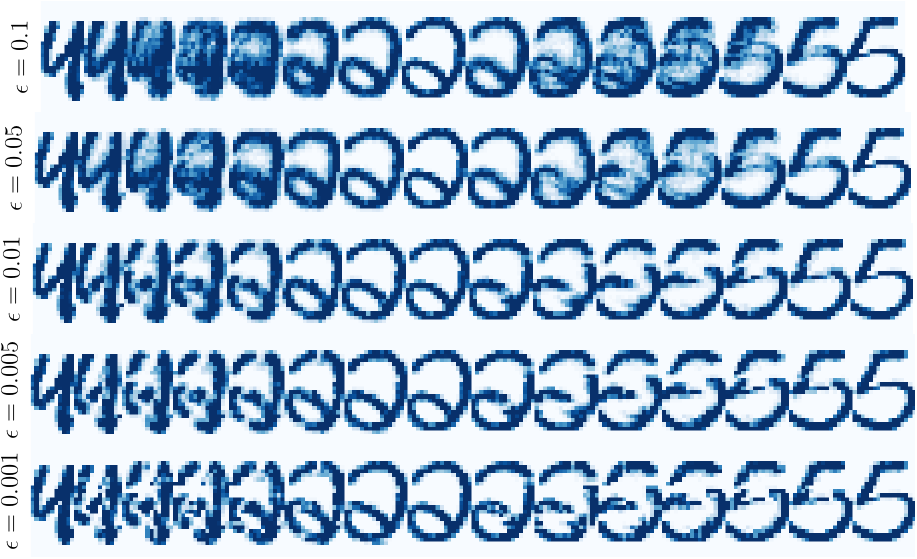}
  \caption{We selected $\epsilon=10^{-2}$ for our MNIST coupling
    experiments as it results in transport maps that are not
    too blurry or sharp.}
  \label{fig:mnist-epsilon}
\end{figure}

\section{Additional experimental and implementation details}
\label{app:exp-details}

Our Jax source code is available at \url{http://github.com/facebookresearch/meta-ot}
and contains:

\newcommand{\fname}[2]{\texttt{#1} \quad #2}
\begin{forest}for tree={folder, grow'=0, inner sep=0pt,}, [
    [\fname{meta\_ot}{Meta OT Python library code}
      [\fname{conjugate.py}{Exact conjugate solver for the continuous setting}]
      [\fname{data.py}{}]
      [\fname{models.py}{}]
      [\fname{utils.py}{}]
    ]
    [\fname{config}{Hydra configuration for the experiments (containing hyper-parameters)}]
    [\fname{train\_discrete.py}{Train Meta OT models for discrete OT}]
    [\fname{train\_color\_single.py}{Train a single ICNN with W2GN between 2 images (for debugging)}]
    [\fname{train\_color\_meta.py}{Train a Meta ICNN with W2GN}]
    [\fname{plot\_mnist.py}{Visualize the MNIST couplings}]
    [\fname{plot\_world\_pair.py}{Visualize the spherical couplings}]
    [\fname{eval\_color.py}{Evaluate the Meta ICNN in the continuous setting}]
    [\fname{eval\_discrete.py}{Evaluate the Meta ICNN for the discrete tasks}]
  ]
\end{forest}

Connecting to the data is one difficulty in running the experiments.
The easiest experiment to re-run is the MNIST one, which will automatically
download the dataset:

\begin{lstlisting}
./train_discrete.py # Train the model, outputting to <exp_dir>
./eval_discrete.py <exp_dir> # Evaluate the learned models
./plot_mnist.py <exp_dir> # Produce further visualizations
\end{lstlisting}

\subsection{Hyper-parameters}
We briefly summarize the hyper-parameters we used for training,
which we did not extensively tune.
In the discrete setting, we use the same hyper-parameters for the
MNIST and spherical settings.

\begin{minipage}{0.5\textwidth}
\begin{table}[H]
  \caption{Discrete OT hyper-parameters.}
  \centering
  \begin{tabular}{rl}\toprule
    Name & Value \\\midrule
    Batch size & 128 \\
    Number of training iterations & 50000 \\
    MLP Hidden Sizes & [1024, 1024, 1024] \\
    Adam learning rate & 1e-3 \\\bottomrule
  \end{tabular}
\end{table}
\end{minipage}
\begin{minipage}{0.5\textwidth}
\begin{table}[H]
  \caption{Continuous OT hyper-parameters.}
  \centering
  \begin{tabular}{rl}\toprule
    Name & Value \\\midrule
    Meta batch size (for $\alpha,\beta$) & 8 \\
    Inner batch size (to estimate $\gL$) & 1024 \\
    Cycle loss weight ($\gamma$) & 3. \\

    Adam learning rate & 1e-3 \\
    $\ell_2$ weight penalty & 1e-6 \\
    Max grad norm (for clipping) & 1. \\

    Number of training iterations & 200000 \\
    Meta ICNN Encoder & ResNet18 \\
    Encoder output size (both measures) & 256$\times$2 \\
    Meta ICNN Decoder Hidden Sizes & [512] \\\bottomrule
  \end{tabular}
\end{table}
\end{minipage}

\subsection{Sinkhorn convergence times, varying thresholds}
In the main paper, \cref{tab:discrete-runtime} reports the runtime of Sinkhorn
to reach a convergence threshold of the marginal error being below a tolerance
of $10^{23}$.
\Cref{tab:all-sinkhorn-runtimes-mnist,tab:all-sinkhorn-runtimes-sphere}
report the results from sweeping over other thresholds and show that
Meta OT's initialization is consistently able to help.

\begin{table}[h]
\caption{Sinkhorn runtime to reach a thresholded marginal error on MNIST.}
\label{tab:all-sinkhorn-runtimes-mnist}
\newcommand{\pair}[2]{$#1$ {\color{gray}\footnotesize $#2$}}
\centering
\begin{tabular}{r|cccc}\toprule
Initialization & Threshold=$10^{-2}$ & Threshold=$10^{-3}$ & Threshold=$10^{-4}$ & Threshold=$10^{-5}$ \\ \midrule
Zeros & \pair{4.5\cdot10^{-3}}{1.5\cdot10^{-3}} & \pair{7.7\cdot10^{-3}}{1.2\cdot10^{-3}} & \pair{1.1\cdot10^{-2}}{1.8\cdot10^{-3}} & \pair{1.5\cdot10^{-2}}{2.3\cdot10^{-3}} \\
Gaussian & \pair{4.1\cdot10^{-3}}{1.2\cdot10^{-3}} & \pair{7.7\cdot10^{-3}}{1.4\cdot10^{-3}} & \pair{1.1\cdot10^{-2}}{1.7\cdot10^{-3}} & \pair{1.4\cdot10^{-2}}{2.4\cdot10^{-3}} \\
Meta OT & \cellhi \pair{2.3\cdot10^{-3}}{9.2\cdot10^{-6}} & \cellhi \pair{3.9\cdot10^{-3}}{1.6\cdot10^{-3}} & \cellhi \pair{6.7\cdot10^{-3}}{1.4\cdot10^{-3}} & \cellhi \pair{1.0\cdot10^{-2}}{2.4\cdot10^{-3}} \\ \bottomrule
\end{tabular}
\end{table}

\begin{table}[h]
\caption{Sinkhorn runtime to reach a thresholded marginal error on the spherical transport problem.}
\label{tab:all-sinkhorn-runtimes-sphere}
\newcommand{\pair}[2]{$#1$ {\color{gray}\footnotesize $\pm #2$}}
\centering
\begin{tabular}{r|cccc}\toprule
Initialization & Threshold=$10^{-2}$ & Threshold=$10^{-3}$ & Threshold=$10^{-4}$ & Threshold=$10^{-5}$ \\ \midrule
Zeros & \pair{8.8\cdot10^{-1}}{1.3\cdot10^{-1}} & \pair{1.4}{1.9\cdot10^{-1}} & \pair{2.1}{3.6\cdot10^{-1}} & \pair{2.8}{5.6\cdot10^{-1}} \\
Gaussian & \pair{5.6\cdot10^{-1}}{9.9\cdot10^{-2}} & \pair{1.1}{2.0\cdot10^{-1}} & \pair{1.7}{3.5\cdot10^{-1}} & \pair{2.4}{5.4\cdot10^{-1}} \\
Meta OT & \cellhi \pair{7.8\cdot10^{-2}}{3.4\cdot10^{-2}} & \cellhi \pair{0.44}{1.5\cdot10^{-1}} & \cellhi \pair{0.97}{3.2\cdot10^{-1}} & \cellhi \pair{1.7}{6.8\cdot10^{-1}} \\ \bottomrule
\end{tabular}
\end{table}

\vfill

\newpage
\subsection{Experimental runtimes and convergence}
\label{app:runtimes}
\begin{figure}[H]
    \centering
    \includegraphics[width=.49\linewidth]{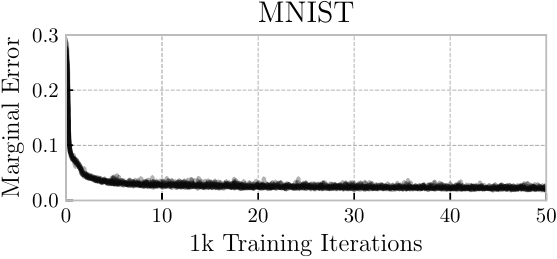}
    \includegraphics[width=.49\linewidth]{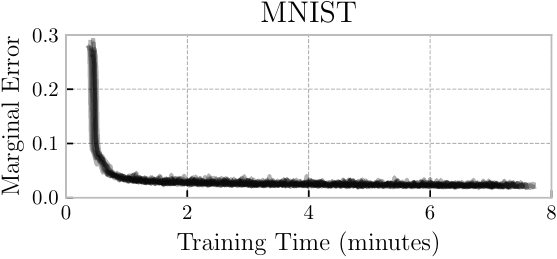} \\[6mm]
    \includegraphics[width=.49\linewidth]{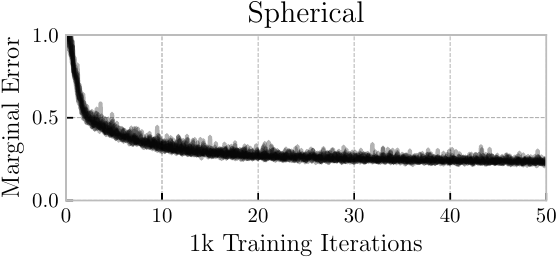}
    \includegraphics[width=.49\linewidth]{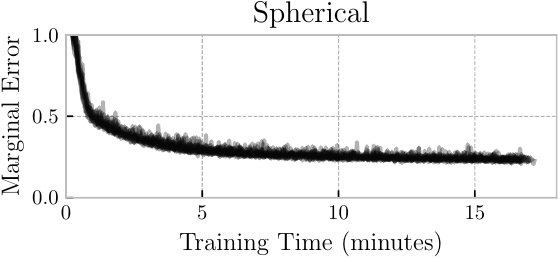} \\[6mm]
    \includegraphics[width=.49\linewidth]{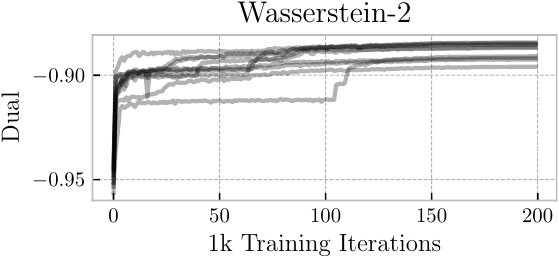}
    \includegraphics[width=.49\linewidth]{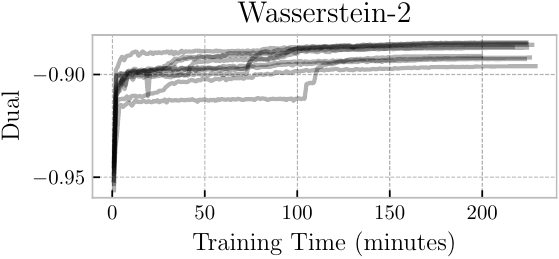}
    \label{fig:training_info}
    \caption{Convergence of Meta OT models during training, reported over
      iterations and wall-clock time. We run each experiment for 10 trials
      with different seeds and report each trial as a line.}
\end{figure}

\Cref{fig:training_info} shows the convergence during training of
Meta OT models in the discrete and continuous settings over 10 trials
on our single Quadro GP100 GPU.
The MNIST models are consistently trained to optimality within 2 minutes (!)
while the continuous model takes a few hours to train.

\newpage
\section{Cross-domain experimental results}
\begin{figure}[H]
    \centering
    \hspace*{-6cm}
    \resizebox{.7\textwidth}{!}{
    \begin{minipage}{5.5in}
    \begin{tikzpicture}[node/.style={inner sep=0,outer sep=0}]
      \node (14) {\includegraphics[width=.39\linewidth]{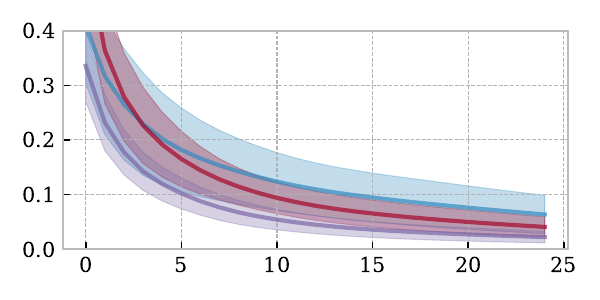}};
      \draw[draw=white,fill=white] ($(14.south west)+(1.2mm,4mm)$) rectangle ++(0.5cm,.9in);
      \node[left=-11mm of 14] (13) {\includegraphics[width=.39\linewidth]{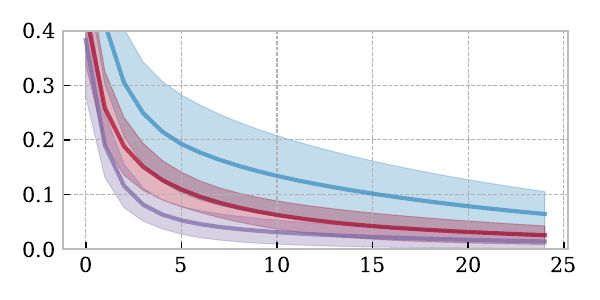}};
      \draw[draw=white,fill=white] ($(13.south west)+(1.2mm,4mm)$) rectangle ++(0.5cm,.9in);
      \node[left=-11mm of 13] (12) {\includegraphics[width=.39\linewidth]{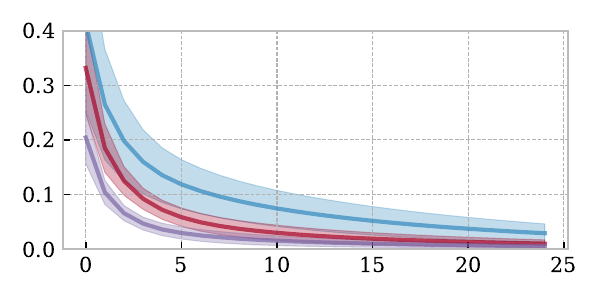}};
      \draw[draw=white,fill=white] ($(12.south west)+(1.2mm,4mm)$) rectangle ++(0.5cm,.9in);
      \node[left=-11mm of 12] (11) {\includegraphics[width=.39\linewidth]{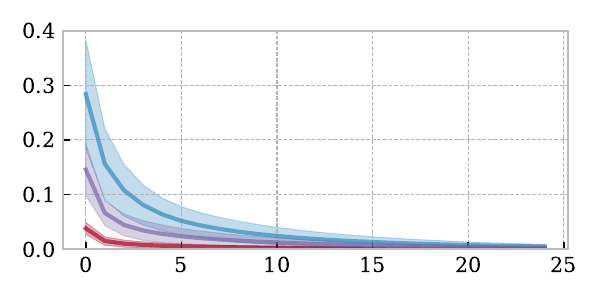}};
      \draw[draw=white,fill=white] ($(11.south west)+(7mm,.5mm)$) rectangle ++(7.3in,5mm);

      \node[below=-6mm of 14] (24) {\includegraphics[width=.39\linewidth]{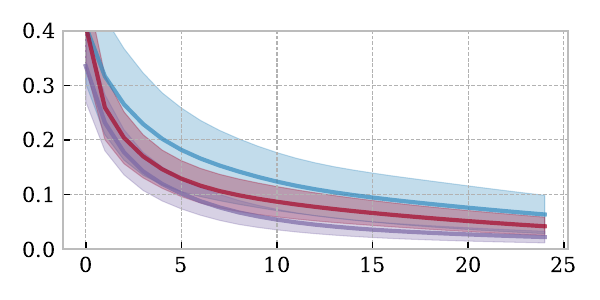}};
      \draw[draw=white,fill=white] ($(24.south west)+(1.2mm,4mm)$) rectangle ++(0.5cm,.9in);
      \node[left=-11mm of 24] (23) {\includegraphics[width=.39\linewidth]{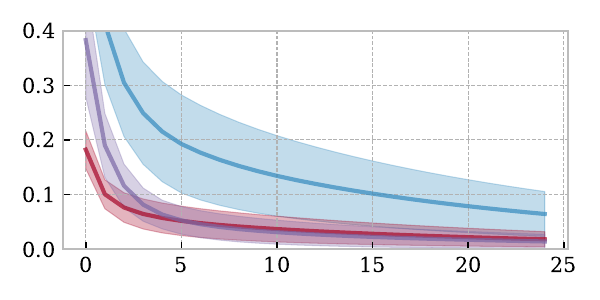}};
      \draw[draw=white,fill=white] ($(23.south west)+(1.2mm,4mm)$) rectangle ++(0.5cm,.9in);
      \node[left=-11mm of 23] (22) {\includegraphics[width=.39\linewidth]{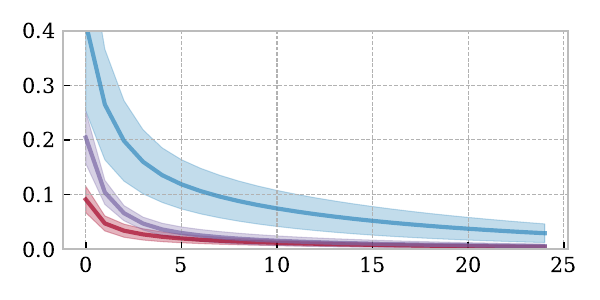}};
      \draw[draw=white,fill=white] ($(22.south west)+(1.2mm,4mm)$) rectangle ++(0.5cm,.9in);
      \node[left=-11mm of 22] (21) {\includegraphics[width=.39\linewidth]{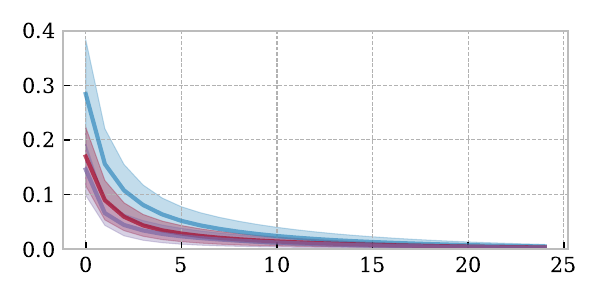}};
      \draw[draw=white,fill=white] ($(21.south west)+(7mm,.5mm)$) rectangle ++(7.3in,5mm);

      \node[below=-6mm of 24] (34) {\includegraphics[width=.39\linewidth]{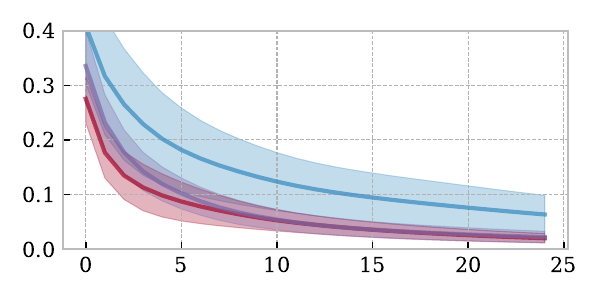}};
      \draw[draw=white,fill=white] ($(34.south west)+(1.2mm,4mm)$) rectangle ++(0.5cm,.9in);
      \node[left=-11mm of 34] (33) {\includegraphics[width=.39\linewidth]{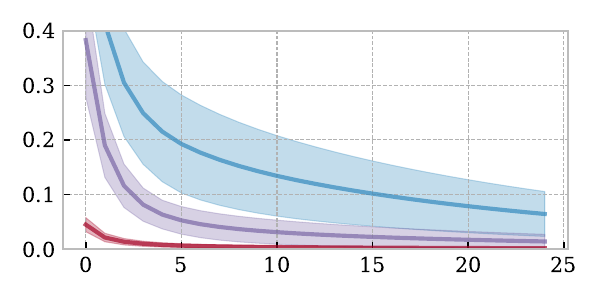}};
      \draw[draw=white,fill=white] ($(33.south west)+(1.2mm,4mm)$) rectangle ++(0.5cm,.9in);
      \node[left=-11mm of 33] (32) {\includegraphics[width=.39\linewidth]{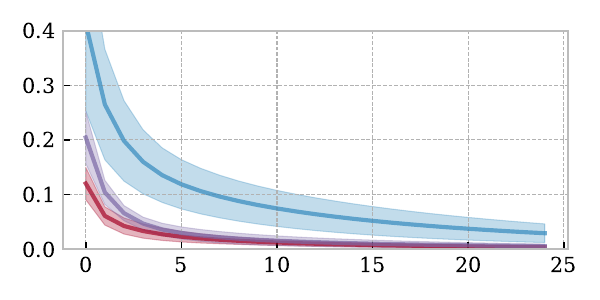}};
      \draw[draw=white,fill=white] ($(32.south west)+(1.2mm,4mm)$) rectangle ++(0.5cm,.9in);
      \node[left=-11mm of 32] (31) {\includegraphics[width=.39\linewidth]{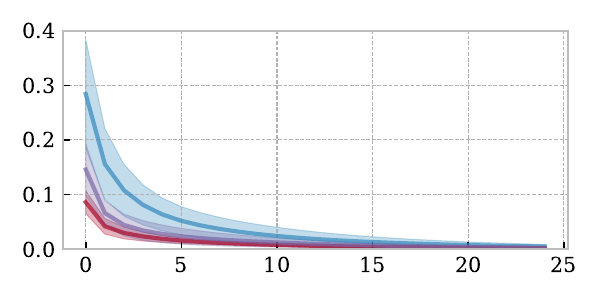}};
      \draw[draw=white,fill=white] ($(31.south west)+(7mm,.5mm)$) rectangle ++(7.3in,5mm);

      \node[below=-6mm of 34] (44) {\includegraphics[width=.39\linewidth]{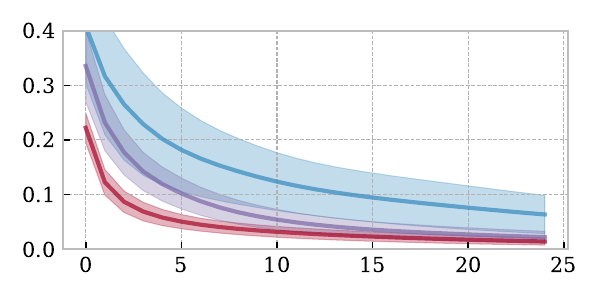}};
      \draw[draw=white,fill=white] ($(44.south west)+(1.2mm,4mm)$) rectangle ++(0.5cm,.9in);
      \node[left=-11mm of 44] (43) {\includegraphics[width=.39\linewidth]{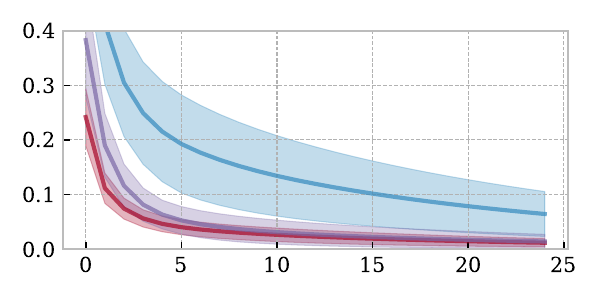}};
      \draw[draw=white,fill=white] ($(43.south west)+(1.2mm,4mm)$) rectangle ++(0.5cm,.9in);
      \node[left=-11mm of 43] (42) {\includegraphics[width=.39\linewidth]{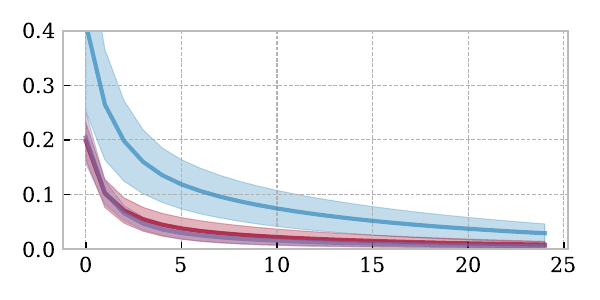}};
      \draw[draw=white,fill=white] ($(42.south west)+(1.2mm,4mm)$) rectangle ++(0.5cm,.9in);
      \node[left=-11mm of 42] (41) {\includegraphics[width=.39\linewidth]{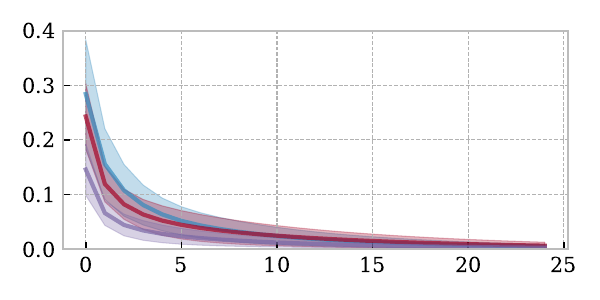}};
      \node [draw=white, below left=-4mm and -4.3cm of 41] () {\footnotesize Sinkhorn Iterations};
      \node [draw=white, below left=-4mm and -4.3cm of 42] () {\footnotesize Sinkhorn Iterations};
      \node [draw=white, below left=-4mm and -4.3cm of 43] () {\footnotesize Sinkhorn Iterations};
      \node [draw=white, below left=-4mm and -4.3cm of 44] () {\footnotesize Sinkhorn Iterations};

      \node [draw=white, rotate=90, below left=-23mm and 3mm of 11] (mnist_label) {MNIST};
      \node [draw=white, rotate=90, below left=-23mm and 3mm of 21] (doodles_label) {Doodles};
      \node [draw=white, rotate=90, below left=-23mm and 3mm of 31] () {USPS};
      \node [draw=white, rotate=90, below left=-23mm and 3mm of 41] () {Uniform};
      \node [draw, rotate=90, below left=-7mm and 13mm of doodles_label] {Training Domain};

      \node [draw=white, above right=-3mm and -3.3cm of 11] () {MNIST};
      \node [draw=white, above right=-3mm and -3.3cm of 12] (doodles_label_eval) {Doodles};
      \node [draw=white, above right=-3mm and -3.3cm of 13] () {USPS};
      \node [draw=white, above right=-3mm and -3.3cm of 14] () {Uniform};
      \node [draw, above right=2mm and 2mm of doodles_label_eval] {Evaluation Domain};

      \draw[draw=black,fill=none,line width=1mm] ($(11.south west)+(7.5mm,5.3mm)$) rectangle ++(45mm,21mm);
      \draw[draw=black,fill=none,line width=1mm] ($(22.south west)+(7.5mm,5.3mm)$) rectangle ++(45mm,21mm);
      \draw[draw=black,fill=none,line width=1mm] ($(33.south west)+(7.5mm,5.3mm)$) rectangle ++(45mm,21mm);
      \draw[draw=black,fill=none,line width=1mm] ($(44.south west)+(7.5mm,5.8mm)$) rectangle ++(45mm,20mm);
    \end{tikzpicture}
    \end{minipage}} \\[2mm]
    Initialization (\cblock{52}{138}{189} Zeros \hspace{2mm}
    \cblock{122}{104}{166} Gaussian \citep{thornton2022rethinking}  \hspace{2mm}
    \cblock{166}{6}{40} Meta OT)
    \label{fig:cross_domain}
    \caption{Cross-domain experiments evaluating how well
      a model trained on one dataset generalizes to another dataset.
      Notably, we are able to train only on a uniform distribution and
      transfer reasonable initializations to the image datasets.
      This indicates that training larger-scale Meta OT models
      for more general classes of discrete OT problems may
      be able to provide a fast and reasonable initialization.
    }
\end{figure}

\section{More information: Meta OT between continuous measures}
\label{sec:appendix:continuous}

\subsection{Meta ICNN Diagram}
\begin{figure*}[h]
  \centering
  \begin{tikzpicture}[every path/.style={thick}]
    \node (im1) {\includegraphics[width=.5in]{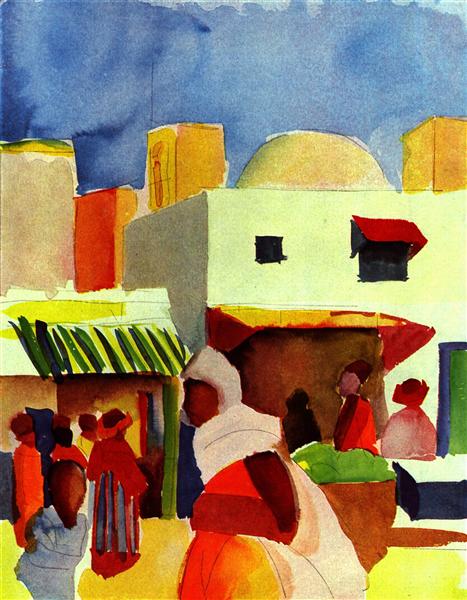}};
    \node[below=0 of im1] (im2) {\includegraphics[width=.5in]{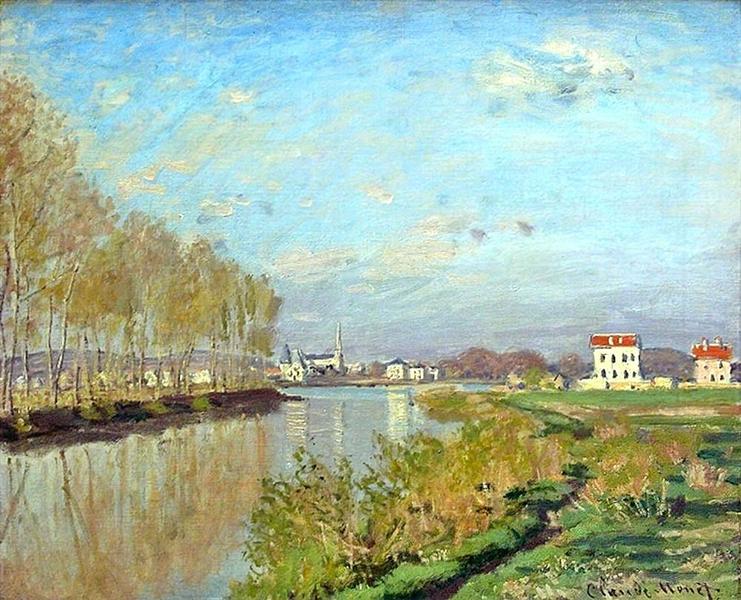}};
    \node[left=0 of im1] (alpha) {$\alpha$};
    \node[left=0 of im2] (beta) {$\beta$};
    \node[right=2cm of im1] (z1) {$z_1$};
    \node[right=2cm of im2] (z2) {$z_2$};
    \node[above=4.5mm of z1] () {$z$};
    \node[xshift=2.5mm] (zmid) at (barycentric cs:z1=1,z2=1) {};
    \path[draw,-] ($(zmid)+(.2mm,0)$) -- ($(zmid)+(-5mm,0)$);
    \scoped[on background layer]{
        \draw[fill=violet!20] ($(zmid) + (-5mm,-15mm)$) rectangle ($(zmid) + (0mm,+15mm)$);
    }
    \node[fill=green!20,draw=black,right=1.5cm of zmid] (varphi) {$\hat \varphi_\theta$};
    \node[below=0mm of varphi] () {Parameters};
    \node[fill=green!20,draw=black,right=1.5cm of varphi] (icnn) {$\psi_{\hat \varphi_\theta}$};
    \node[below=0mm of icnn] () {ICNN};
    \node[fill=red!20,draw=black,right=1.5cm of icnn] (icnn_grad) {$\hat T(\cdot)=\nabla_x\psi_{\hat \varphi_\theta}(\cdot)$};
    \node[below=0mm of icnn_grad] () {Transport map};
    \path[draw,->] (im1) -- node[above]{ResNet$_\theta$} (z1);
    \path[draw,->] (im2) -- node[above]{ResNet$_\theta$} (z2);
    \path[draw,->] (zmid) -- node[above]{MLP$_\theta$} (varphi);
    \edge[] {varphi} {icnn};
    \edge[] {icnn} {icnn_grad};
  \end{tikzpicture} \\[-1.5mm]
  \caption{A Meta ICNN for image-based input measures.
    A shared ResNet processes the input measures
    $\alpha$ and $\beta$ into latents $z$ that are decoded
    with an MLP into the parameters $\varphi$ of an ICNN dual
    potential $\psi_\varphi$.
    The derivative of the ICNN provides the transport map $\hat T$.
  }
  \label{fig:meta-icnn}
\end{figure*}
\subsection{Other models for continuous OT}
\label{app:other-W2-models}
We explored a hyper-network model because it is conceptually the
most similar to predicting the optimal dual variables in the continuous
setting and results in rapid predictions.
However, it may not scale well to predicting high-dimensional
parameters of ICNNs. This section presents two alternatives
based on MAML \citep{pmlr-v70-finn17a} and
neural processes \citep{garnelo2018neural,garnelo2018conditional},
and conditional OT maps \citep{bunne2022supervised}.

\subsubsection{Optimization-based meta-learning (MAML-inspired)}
The model-agnostic meta-learning setup proposed in MAML \citep{pmlr-v70-finn17a}
could also be applied in the Meta OT setting to learn an adaptable
initial parameterization.
In the continuous setting, one initial version would take a parameterized
dual potential model $\psi_\varphi(x)$ and seek to learn an initial
parameterization $\varphi_0$ so that optimizing a loss such as the
W2GN loss $\gL$ from \cref{eq:w2gn-loss} results in a minimal
$\gL(\varphi_K)$ after adapting the model for $K$ steps.
Formally, this would optimize:
\begin{equation}
  \argmin_{\varphi_0} \gL(\varphi_K)\quad \text{where}\quad \varphi_{t+1}=\varphi_t-\nabla_\varphi\gL(\varphi_t)
  \label{eq:maml-meta-ot}
\end{equation}

\citet{tancik2021learned} explores similar learned initializations
for coordinate-based neural implicit representations for 2D images,
CT scan reconstruction, and 3d shape and scene recovery from 2D observations.

\textbf{Challenges for Meta OT.}
The transport maps given by $T=\nabla\psi$ can significantly
vary depending on the input measures $\alpha,\beta$.
We found it difficult to learn an initialization that can be rapidly
adapted, and optimizing \cref{eq:maml-meta-ot} is
more computationally expensive than
\cref{eq:amor-w2gn-loss} as it requires unrolling through many
evaluations of the transport loss $\gL$.
And, we found that \emph{only} learning to predict the optimal
parameters with \cref{eq:amor-w2gn-loss}, conditional on the input measures,
and then fine-tuning with W2GN to be stable.

\textbf{Advantages for Meta OT.}
Exploring MAML-inspired methods could further incorporate the knowledge
that the model's prediction is going to be fine-tuned into the
learning process.
One promising direction we did not try could be to integrate
some of the ideas from
LEO \citep{rusu2018meta} and CAVIA \citep{zintgraf2019fast},
which propose to learn a latent space for the parameters
where the initialization is also conditional on the input.

\subsubsection{Neural process and conditional Monge maps}
The (conditional) neural process models considered in \citet{garnelo2018neural,garnelo2018conditional}
can also be adapted for the Meta OT setting, and is similar to the
model proposed in \citet{bunne2022supervised}.
In the continuous setting, this would result in a
dual potential that is also conditioned on a representation
of the input measures, \eg $\psi_\varphi(x; z)$ where
$z\defeq f^{\rm emb}_\varphi(\alpha, \beta)$ is a learned embedding
of the input measures that is learned with the parameters of $\psi$.
This could be formulated as
\begin{equation}
  \argmin_\varphi \E_{(\alpha,\beta)\sim\gD} \gL(\varphi, f^{\rm emb}_\varphi(\alpha, \beta)),
  \label{eq:neural-process-meta-ot}
\end{equation}
where $\gL$ modifies the model used in the loss \cref{eq:w2gn-loss} to
also be conditioned on the context extracted from the measures.

\textbf{Challenges for Meta OT.}
This raises the issue on best-formulating the model to be conditional
on the context. One way could be to append $z$ to the input
point $x$ in the domain.
\citet{bunne2022supervised} proposes to use the Partially Input-Convex
Neural Network (PICNN) from \citep{amos2017input} to make the model
convex with respect to $x$ and not $z$.

\textbf{Advantages for Meta OT.}
A large advantage is that the representation $z$ of the measures
$\alpha,\beta$ would be significantly lower-dimensional than the
parameters $\varphi$ that our Meta OT models are predicting.

\newpage
\subsection{Continuous Wasserstein-2 color transfer}
\label{app:color}

The following public domain images are from
\href{https://www.wikiart.org/}{WikiArt}:

\begin{itemize}
\item \href{https://www.wikiart.org/en/winston-churchill/distant-view-of-the-pyramids-1921}{Distant View of the Pyramids by Winston Churchill (1921)}
\item \href{https://www.wikiart.org/en/claude-monet/charing-cross-bridge-overcast-weather}{Charing Cross Bridge, Overcast Weather by Claude Monet (1900)}
\item \href{https://www.wikiart.org/en/claude-monet/houses-of-parliament}{Houses of Parliament by Claude Monet (1904)}
\item \href{https://www.wikiart.org/en/childe-hassam/october-sundown-newport-1901}{October Sundown, Newport by Childe Hassam (1901)}
\item \href{https://www.wikiart.org/en/juan-gris/landscape-with-house-at-ceret-1913}{Landscape with House at Ceret by Juan Gris (1913)}
\item \href{https://www.wikiart.org/en/claude-monet/irises-in-monet-s-garden-03}{Irises in Monet's Garden by Claude Monet (1900)}
\item \href{https://www.wikiart.org/en/paul-klee/crystal-1921}{Crystal Gradation by Paul Klee (1921)}
\item \href{https://www.wikiart.org/en/paul-klee/senecio-1922}{Senecio by Paul Klee (1922)}
\item \href{https://www.wikiart.org/en/josef-capek/vaza-s-kvetinami-1914}{Váza s květinami by Josef Capek (1914)}
\item \href{https://www.wikiart.org/en/vincent-van-gogh/sower-with-setting-sun-1888-3}{Sower with Setting Sun by Vincent van Gogh (1888)}
\item \href{https://www.wikiart.org/en/claude-monet/three-trees-in-grey-weather}{Three Trees in Grey Weather by Claude Monet (1891)}
\item \href{https://www.wikiart.org/en/vincent-van-gogh/vase-with-daisies-and-anemones-1887}{Vase with Daisies and Anemones by Vincent van Gogh (1887)}
\end{itemize}
\newpage

\def\photos{0004,0006,0007,0009,0010,0012}
\def\head{{\Large $\alpha$ \hspace{1.1in} $\beta$ \hspace{1.1in} $T_\#\alpha$ \hspace{.9in} $T_\#^{-1}\beta$}}
\begin{figure}[H]
  \vspace{-2mm}
  \centering
  \head \\
  \begin{minipage}{0.8\linewidth}
  \foreach \I in \photos {\includegraphics[width=\textwidth]{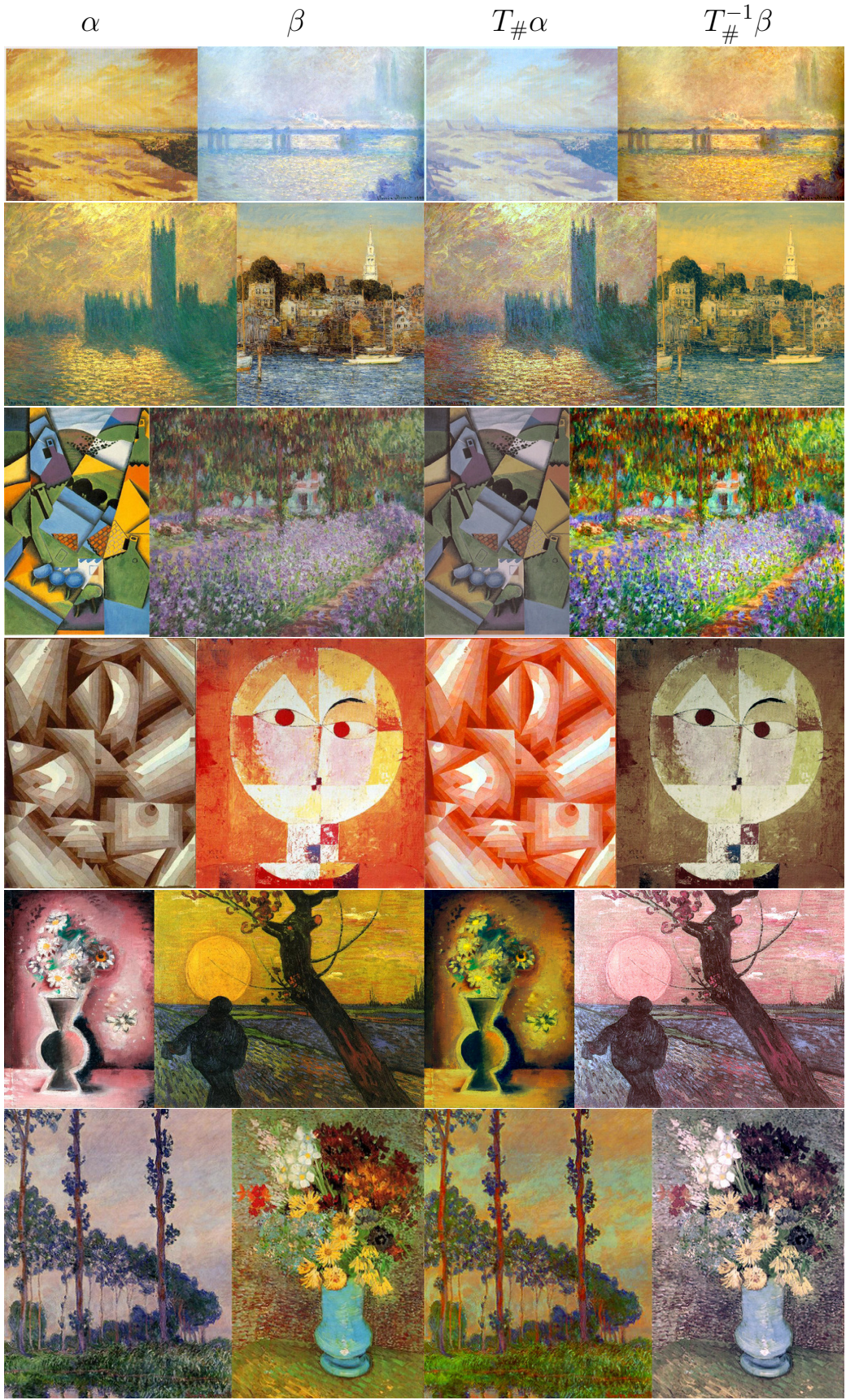} \\}
  \end{minipage}
  \vspace*{-5mm}
  \caption{Meta ICNN (initial prediction).
    The sources are given in the beginning of \cref{app:color}.
  }
\end{figure}

\newpage

\begin{figure}[H]
  \vspace{-2mm}
  \centering
  \head \\
  \begin{minipage}{0.8\linewidth}
  \foreach \I in \photos {\includegraphics[width=\textwidth]{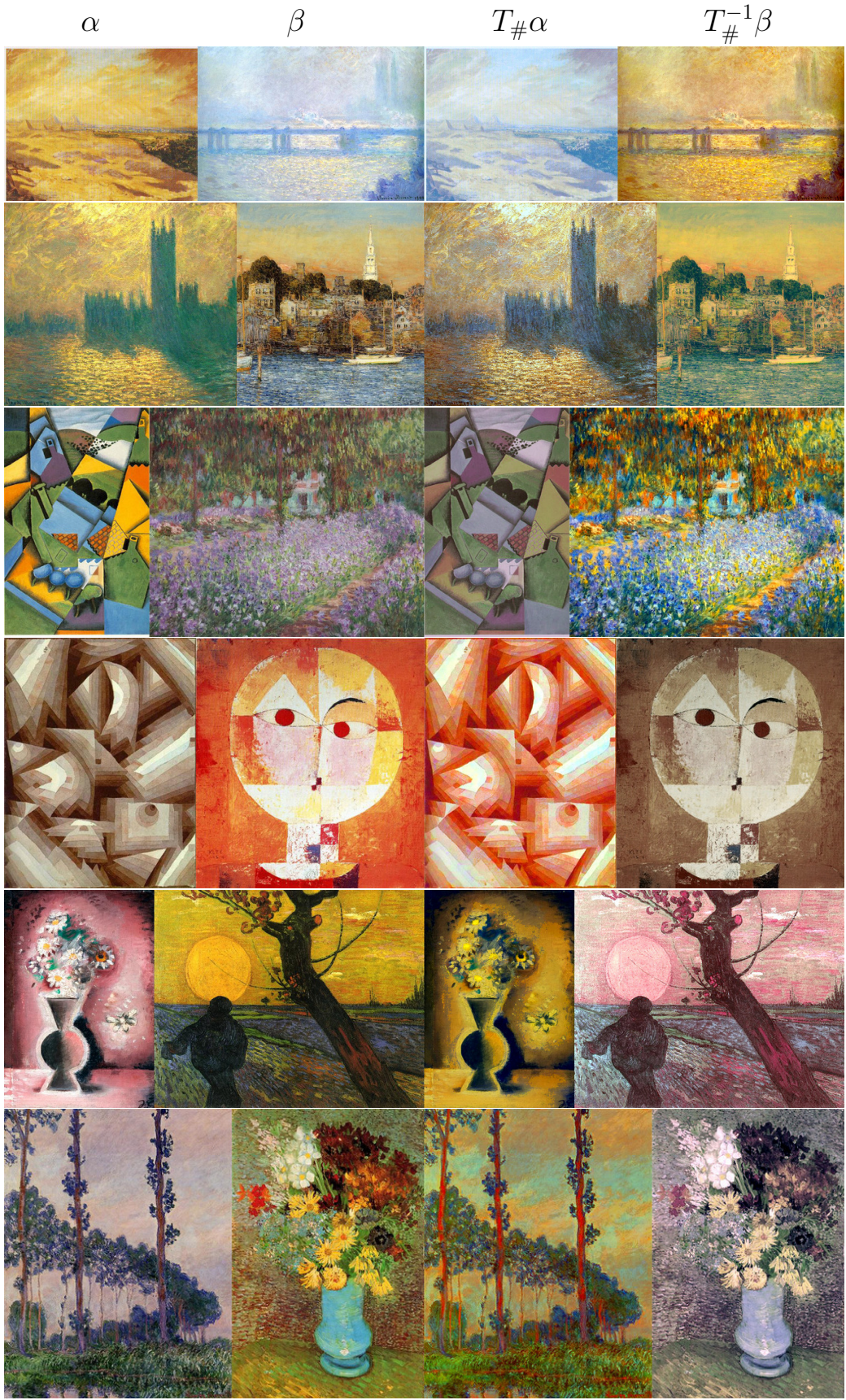} \\}
  \end{minipage}
  \vspace*{-5mm}
  \caption{Meta ICNN + W2GN fine-tuning.
    The sources are given in the beginning of \cref{app:color}.}
\end{figure}

\newpage
\begin{figure}[H]
  \vspace{-2mm}
  \centering
  \head \\
  \begin{minipage}{0.8\linewidth}
  \foreach \I in \photos {\includegraphics[width=\textwidth]{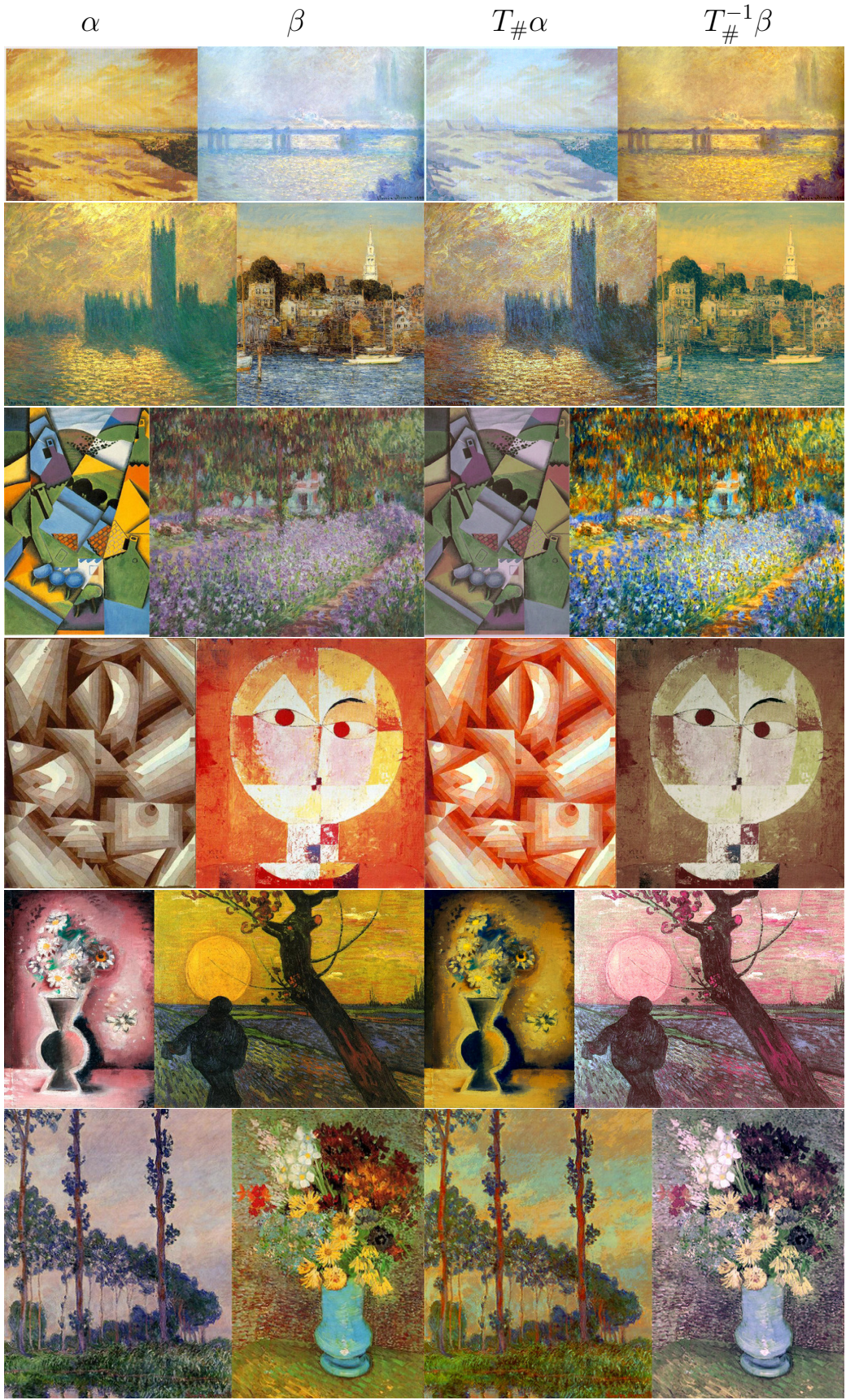} \\}
  \end{minipage}
  \vspace*{-5mm}
  \caption{W2GN (final). The sources are given in the beginning of \cref{app:color}.}
\end{figure}

\end{document}